\documentclass[letterpaper, 10 pt, conference]{ieeeconf}  % Comment this line out if you need a4paper

\IEEEoverridecommandlockouts                              % This command is only needed if 
\usepackage{xcolor}
\usepackage[cmex10]{amsmath}
\usepackage{amssymb}
\usepackage{graphicx}
\usepackage{comment,xspace}
\usepackage[bookmarks=true]{hyperref}
\usepackage{booktabs}
\usepackage{algorithm,algorithmic} 
\usepackage{fancybox}
\usepackage{times}
\usepackage{multicol}
\usepackage{enumerate}
\usepackage{bm}                % bold math
\usepackage{xfrac}
\usepackage{epstopdf}

\makeatletter
\let\NAT@parse\undefined
\makeatother
\usepackage[numbers]{natbib}	% numbers option provides compact numerical references in the text. 

\newcommand{\real}{{\mathbb{R}}}
\newcommand{\reals}{\real}

\newcommand{\prob}[1]{\mbox{$\mathbb{P}\left(#1\right)$}} 
 
\newcommand{\probcond}[2]{\mbox{$\mathbb{P}\left(#1 \,\mid \, #2\right)$}}

\newcommand{\suchthat}{\;\ifnum\currentgrouptype=16 \middle\fi|\;}  

\usepackage{amssymb}
\usepackage{graphicx}
\usepackage{xspace}
\usepackage{caption}
\usepackage{subcaption}
\usepackage{algorithmic} 
\usepackage{algorithm}

\graphicspath{{./fig/}}

\newcommand{\pump}{PUMP\xspace}
\newcommand{\skp}{SKP\xspace}

\newcommand{\mcmp}{MCMP\xspace}
\newcommand{\hsmc}{HSMC\xspace}
\usepackage{multicol}
\usepackage{color}

\newcommand{\dr}{\lambda}   % ratio of \delta to r, i.e. \delta = \dr r

\newcommand{\Popen}{P_{\mathrm{open}}}

\newcommand{\Psoln}{P_{\xgoal}}
\newcommand{\B}{\mathcal{G}}   % \mathcal{G} for ``group''
\newcommand{\cptarget}{\alpha}

\newcommand{\xfree}{\mathcal X_{\text{free}}}
\newcommand{\xobs}{\mathcal X_{\text{obs}}}
\newcommand{\xgoal}{\mathcal X_{\text{goal}}}
\newcommand{\xinit}{x_{\mathrm{init}}}

\newcommand{\union}{\cup}

\newcommand{\N}{\mathbb{N}}

\newcommand{\x}{\mathbf{x}}
\renewcommand{\u}{\mathbf{u}}
\newcommand{\y}{\mathbf{y}}
\renewcommand{\a}{\mathbf{a}}
\renewcommand{\b}{\mathbf{b}}
\renewcommand{\d}{\mathbf{d}}
\newcommand{\dx}{\delta\mathbf{x}}
\newcommand{\du}{\delta\mathbf{u}}
\newcommand{\dy}{\delta\mathbf{y}}
\newcommand{\xn}{\mathbf{x}^{\text{nom}}} 
\newcommand{\un}{\mathbf{u}^{\text{nom}}}
\newcommand{\yn}{\mathbf{y}^{\text{nom}}}

\newcommand{\z}{\mathbf{z}}

\renewcommand{\v}{\text{\bf{v}}}

\newcommand{\w}{\mathbf{w}}
\newcommand{\dt}{{\Delta t}}

\newcommand{\esmargin}[2]{#1}

\DefineNamedColor{named}{darkgreen}{rgb}{0,0.8,0}
\newcommand{\rev}[1]{{\color[named]{black}#1}}%\marginpar{\color[named]{darkgreen}\Large \bf $+$}}
\newcommand{\revv}[1]{{\color[named]{black}#1}}%\marginpar{\color[named]{darkgreen}\Large \bf $+$}}

\usepackage{eqparbox}

\newcommand{\LINEIFELSE}[3]{%
    \STATE\algorithmicif\ {#1}\ \algorithmicthen\ {#2} \algorithmicelse\ {#3} \algorithmicend\ \algorithmicif }
\newcommand{\LINEIF}[2]{%
    \STATE\algorithmicif\ {#1}\ \algorithmicthen\ {#2} \algorithmicend\ \algorithmicif }

\usepackage[font={small}]{caption}

 \renewcommand{\baselinestretch}{0.92}

\title{\LARGE \bf
Real-Time Stochastic Kinodynamic Motion Planning \\
via Multiobjective Search on GPUs}

\author{Brian Ichter, Edward Schmerling, Ali-akbar Agha-mohammadi, and Marco Pavone% <-this % stops a space
\thanks{Brian Ichter and Marco Pavone are with the Department of Aeronautics and Astronautics, Stanford University. Edward Schmerling is with the Institute for Computational and Mathematical Engineering, Stanford University, Stanford, CA 94305. Ali-akbar Agha-mohammadi is with the Jet Propulsion Laboratory, California Institute of Technology, Pasadena, CA 91109. 
        {\tt\footnotesize \{ichter, schmrlng, pavone\}@stanford.edu, aliakbar.aghamohammadi@jpl.nasa.gov}%
 }%
\thanks{This work was supported by a Qualcomm Innovation Fellowship and by NASA under the Space Technology Research Grants Program, Grant NNX12AQ43G. Brian Ichter was supported by the DoD NDSEG Program.}% <-this % stops a space
}

\begin{document}

\maketitle
\thispagestyle{empty}
\pagestyle{empty}

\begin{abstract}
In this paper we present the \pump (Parallel Uncertainty-aware Multiobjective Planning) algorithm for addressing the stochastic kinodynamic motion planning problem, whereby one seeks a low-cost, dynamically-feasible motion plan subject to a constraint on collision probability (CP).
\rev{To ensure exhaustive evaluation of candidate motion plans (as needed to tradeoff the competing objectives of performance and safety), \pump incrementally builds the Pareto front of the problem, accounting for the optimization objective {and} an approximation of CP.
This is performed by a massively parallel multiobjective search, here implemented with a focus on GPUs.
Upon termination of the exploration phase, \pump searches the Pareto set of motion plans to identify the lowest cost solution  that is certified to satisfy the CP constraint (according to an asymptotically exact estimator).}
We introduce a novel particle-based CP approximation scheme, designed for efficient GPU implementation, which accounts for dependencies over the history of a trajectory execution. 
We present numerical experiments for quadrotor planning wherein \pump identifies solutions in \texttildelow100 ms, evaluating over one hundred thousand partial plans through the course of its exploration phase. 
The results show that this multiobjective search achieves a lower motion plan cost, for the same CP constraint, compared to a safety buffer-based search heuristic and repeated RRT trials.
\end{abstract}

\section{Introduction}\label{sec:intro}

\begin{figure}[h]
    \centering
    \begin{subfigure}[b]{0.21\textwidth}
        \includegraphics[width=\textwidth]{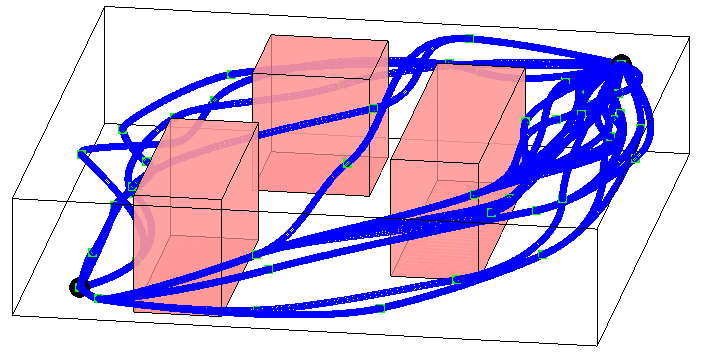}
        \caption{}
        \label{fig:pump_homotopy3obs}
    \end{subfigure}      
    \begin{subfigure}[b]{0.26\textwidth}
        \includegraphics[width=\textwidth]{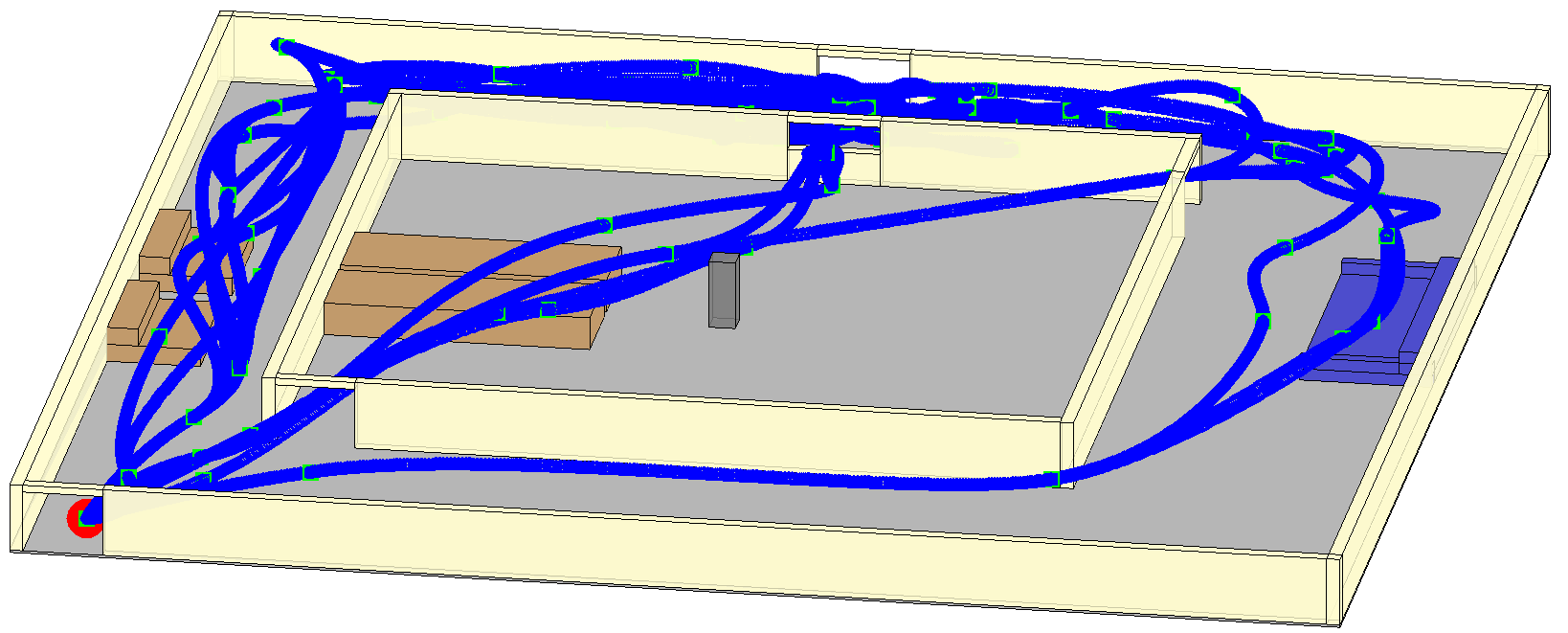}
        \caption{}
        \label{fig:pump_homotopyIndoor}
    \end{subfigure}    
    \begin{subfigure}[b]{0.36\textwidth}
        \includegraphics[width=\textwidth]{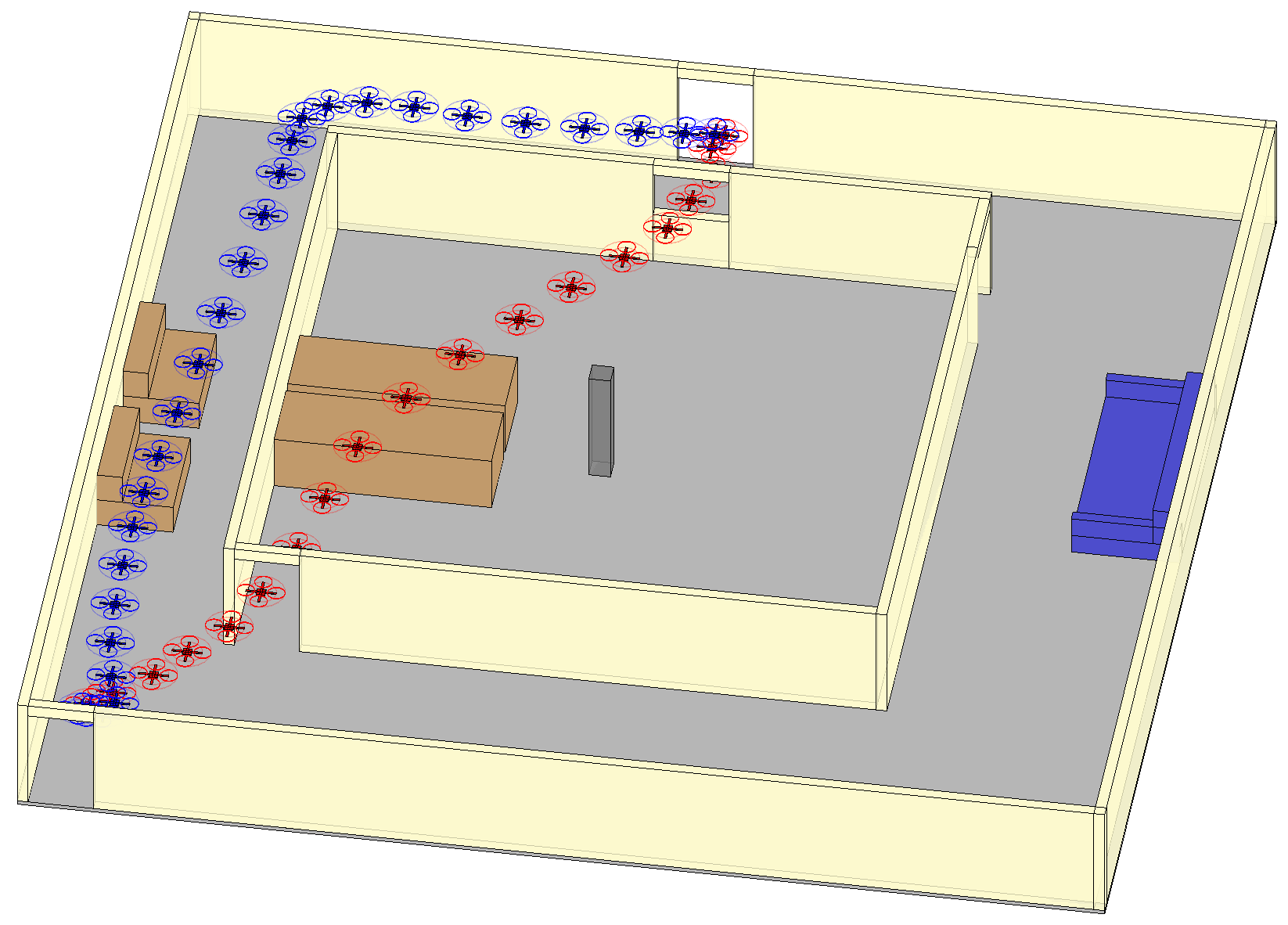}
        \caption{}
        \label{fig:pump_homotopy2paths}
    \end{subfigure}    
\caption{(\ref{fig:pump_homotopy3obs}-\ref{fig:pump_homotopyIndoor}) Illustration of candidate motion plans identified by \pump's exploration phase, representing multiple homotopy classes. (\ref{fig:pump_homotopy2paths}) \rev{Selected} trajectories for a 2\% (blue) and 5\% (red) target CP; the 2\% CP motion plan takes a less aggressive route.}
\label{fig:pump_homotopy}
\end{figure}

Motion planning is a fundamental problem in robotics involving the computation of a trajectory from an initial state to a goal state that avoids obstacles and optimizes an objective function \cite{LaValle2006}. 
This problem has traditionally been considered in the context of deterministic systems without kinematic or dynamic constraints.  \rev{On the other hand, autonomous robotic systems are poised to operate in increasingly dynamic, partially-known environments  where these simplifying assumptions may no longer hold.}
\rev{Successful navigation in these settings requires quick actions that fully exploit the system's dynamical capabilities and are robust to uncertainties in sensing and actuation.}
Proper consideration of trajectory cost should bring trajectory optimization against the limits of safety design constraints; a well designed trajectory should therefore be situated on a Pareto optimal front representing the tradeoff between the competing objectives of performance and safety.

Prompted by these considerations, in this work we study the problem of motion planning under uncertainty. We focus, in particular, on \rev{the stochastic kinodynamic motion planning formulation investigated in \cite{JansonSchmerlingEtAl2015b} (henceforth referred to as the \skp problem),} whereby one seeks a low-cost reference trajectory to be tracked subject to a constraint on obstacle collision probability (CP).
Our key contribution is to introduce the Parallel Uncertainty-aware Multiobjective Planning algorithm, a novel approach to planning under uncertainty that \rev{initially} employs a multiobjective search to build a Pareto front of plans within the state space, actively considering both cost and an approximation of CP.
\rev{As shown in Fig.~\ref{fig:pump_homotopy}}, \rev{this search identifies several high-quality trajectories to be considered as solutions. 
A bisection search is then performed over the Pareto set of solutions to identify the lowest cost reference trajectory that can be tracked with satisfaction of the CP constraint (verified via Monte Carlo (MC) methods \cite{JansonSchmerlingEtAl2015b}).
Such a procedure enables the explicit tradeoff of performance and safety, as opposed to several existing approaches that forgo plan optimization and recast the problem as an unconstrained planning problem where trajectory CP is minimized (with, possibly, a-posteriori heuristics to improve upon solution cost).}
The key to enabling real-time solutions in this work lies in \rev{algorithm parallelization to leverage 
GPU hardware and the development of a novel fast and accurate CP approximation strategy over trajectories.}

\textit{Related Work:} \rev{A variety of strategies have been studied for planning under uncertainty.}
One choice is to formalize the planning problem as a partially observable Markov Decision Process (POMDP) \cite{KaelblingLittmanEtAl1998,KurniawatiHsuEtAl2008,LeeKim2016}. 
This allows a planner to optimize over closed-loop control policies that map belief distributions over the robot state to appropriate actions. Considering POMDP problems in their full generality, however, has been recognized as extremely computationally intensive; \cite{Agha-mohammadiChakravortyEtAl2014} provides a method for remedying the complexity of tracking the evolution of the state belief. In that work the authors use local feedback controllers to regularize the belief state at certain waypoints in space in order to break the dependence on history. While effective for constructing control policies that can handle large uncertainties, \rev{this approach is limited in the context of trajectory optimization due to the required stabilization phases which may adversely affect trajectory cost}. For systems where robustness against large deviations is not a routine need, a common alternative approach is to optimize over open-loop trajectories, producing control sequences which serve as control policies when augmented with a tracking controller. These reference trajectories are recomputed in a receding horizon fashion to address tracking errors and environmental uncertainties.
This is the control strategy advanced in \cite{JansonSchmerlingEtAl2015b,LudersKothariEtAl2010,BergAbbeelEtAl2011,SunPatilEtAl2015}; we note that high-frequency replanning \cite{SunPatilEtAl2015} has been shown to yield \esmargin{greatly increased robustness for systems with significant uncertainty.}{I have (somewhat intentionally for space reasons) left out any more detailed motivation for why we consider real-time operation as opposed to offline roadmaps, however, Marco may deem it necessary}
These works view the problem through the lens of chance-constrained optimization \cite{BlackmoreOnoEtAl2010}, of which the \skp is one formulation, where trajectories are gauged by a metric of their safety in addition to cost.
Computationally efficient methods for the asymptotically exact estimation of CP were introduced in \cite{JansonSchmerlingEtAl2015b};  however, the Monte Carlo motion planning procedure in that work addresses the relationship between cost and CP only indirectly, employing a deterministic asymptotically optimal planner with a safety-buffer heuristic as a proxy for CP. \rev{
The methods in \cite{BergAbbeelEtAl2011,SunPatilEtAl2015} follow an alternative approach towards chance-constraint optimization, whereby one employs parallel rapidly-exploring random trees (RRTs) in a safety-agnostic setting,
to generate a large set of candidate trajectories for CP evaluation; high trajectory quality is ensured only through evaluation of many full-length trajectories. 
In this work we draw inspiration from the multiobjective literature to select reference trajectories
 % for the \skp problem
in a principled, {\em incremental} fashion during exploration to achieve exhaustivity similar to evaluating many full motion plans.
This enables rapid planning without compromising on cost or the ability to guarantee a hard safety constraint.}

Unfortunately, even with a known state space (e.g., given a graph of all possible states and motions), multiobjective search with just two objectives is already NP-hard \cite{SandersMandow2013}. 
\rev{Furthermore, the allocation of risk along a trajectory is a combinatorial problem.
While basic formulations have been solved with computationally intensive optimizations, such as mixed-integer linear programs or constrained nonlinear programs, only brute force techniques are amenable to problems involving dynamics, high-dimensions, or cluttered spaces \cite{LudersKothariEtAl2010}.}
Thus a major tenet of this work is to leverage algorithm parallelization to exploit the computational breadth of GPUs.
An early result in sampling-based motion planning showed that probabilistic roadmap methods are embarrassingly parallel \cite{AmatoDale1999}, which later inspired implementation on GPUs \cite{PanLauterbachEtAl2010}. 
Most recent works in parallelization of sampling-based motion planning have focused on specific subroutines (such as collision checking and nearest neighbor search) \cite{PanLauterbachEtAl2010b,BialkowskiKaramanEtAl2011} or used AND-/OR-parallelism to adapt serial algorithms \cite{SunPatilEtAl2015,DevaursSimeonEtAl2013,ParkPanEtAl2014}. 
 \rev{In our work, the {\em design} of the entire algorithm (as opposed to isolated subroutines) is motivated by massive parallelization, and efficient GPU implementation -- in particular the CP approximation and multiobjective exploration.}

\textit{Statement of Contributions:} In this work we present the Parallel Uncertainty-aware Multiobjective Planning algorithm (\pump), capable of returning high quality solutions to the stochastic kinodynamic motion planning problem. \esmargin{At PUMP's core}{before this sentence is where I would ordinarily state that we're considering an LQG setup, but is it too near the top of the SoC?} is a multiobjective search that considers both trajectory cost and an approximation of collision probability while exploring the state space; in this way, a Pareto front of high-quality motion plans is identified for later certification of CP constraint satisfaction via MC methods. The additional computational burden of maintaining a Pareto front of plans is offset \rev{through algorithm design for massively parallel computation on GPU hardware.}

The design of \pump can be separated into three phases, each of which is well posed for application to GPUs. The first is a graph building phase which constructs a representation of available motions within the free state space, as inspired by embarrassingly parallel probabilistic roadmap methods \cite{AmatoDale1999}. 
The second explores this graph through a multiobjective search, building a Pareto front of motion plans outward from the initial state by expanding in parallel the set of all frontier plans below a constantly increasing cost threshold. This expansion is followed by a dominance check that considers the cost and CP (approximated using a heuristic) of all plans arriving at the same state to aggressively remove those that are less promising.
We note that propagating CP along a motion plan requires consideration of the uncertainty distribution conditioned on collision avoidance; for each partial motion plan we collapse the full shape of the uncertainty distribution to a single approximate CP value for dominance comparison. 
We show that this heuristic works well in practice and is necessary for limiting search branching factor.
Upon completion of this graph exploration phase, a bisection search is performed over the Pareto optimal set of plans reaching the goal according to their approximate CP. The intent of this third phase is to correct for any CP inaccuracy by computing an asymptotically exact estimate of the collision probability using MC-based methods \cite{JansonSchmerlingEtAl2015b} (an embarrassingly parallel process), to certify that the final returned solution satisfies the problem constraint.

Through numerical experiments with a \rev{simplified quadrotor model,} we find that 
PUMP improves over Monte Carlo Motion Planning \cite{JansonSchmerlingEtAl2015b} and repeated RRT trajectory generation in terms of motion plan cost, given the same CP constraint, while returning results in a tempo compatible with real-time operation, on the order of 100 ms.
To achieve this run time, a new trajectory CP approximation method is presented, termed Half-Space Monte Carlo (\hsmc). 
\revv{The method proceeds by sampling many realizations of a reference-tracking controller and checking each against a series of relevant half-spaces representing a simplified collision model, allowing a fast, accurate, and trivially parallelized CP approximation.}

\textit{Organization:} This paper is structured as follows. Section~\ref{sec:problem} \rev{formally defines the stochastic kinodynamic motion planning problem addressed in this work.}
Section~\ref{sec:cpapprox} outlines previous collision probability approximation methods and presents the novel \hsmc method. 
Section~\ref{sec:pump} discusses the \pump algorithm and % that leverages a multiobjective search to determine a Pareto optimal set of plans, ultimately returning the lowest-cost trajectory subject to a CP constraint. 
Section~\ref{sec:sims} describes its implementation and presents simulation results supporting our statements. %, followed by simulations that show real-time performance and provide comparison to a state-of-the-art planning under uncertainty algorithm.
Lastly, Section~\ref{sec:conclusions} discusses conclusions and future work.

\section{Problem Statement}\label{sec:problem}

This paper addresses the stochastic kinodynamic motion planning (\skp) problem \rev{formulation} posed in \cite{JansonSchmerlingEtAl2015b} which we briefly review here.
We consider a robot described by linear dynamics with Gaussian process and measurement noise which tracks a nominal trajectory using a Linear-Quadratic Gaussian (LQG)-derived controller.
\esmargin{Although \pump with Monte Carlo simulation is equipped, as an algorithm framework, to handle problem formulations generalized to a nonlinear setup (as in, e.g., \cite{BergAbbeelEtAl2011,SunPatilEtAl2015} which base their computations on a local linearization around the reference trajectory) and to
any tracking controller (e.g., LQG with nonlinear extended Kalman filter estimation, or geometric tracking control \cite{LeeLeokyEtAl2010}), we focus our attention in this paper on demonstrating improvement in performance and speed over existing LQG methods.}{weak, especially the last clause, but it is what it is}
We note, however, that the Monte Carlo methods discussed in this work --- in particular, all of their intermediate inputs computed from dynamics --- are in principle directly applicable to a local linearization, or even non-Gaussian noise sampled, e.g., from zero-mean stable distributions.

The underlying nominal trajectory (and corresponding feedfoward control term) is planned assuming continuous dynamics, but we assume discretized \esmargin{(zero-order hold)}{i admit we're getting heavy on the parentheticals at this point} approximate dynamics for the tracking controller. That is, the dynamics evolve according to the stochastic linear model:
\begin{equation}
\label{eq:cont}
\small
\dot \x(t) = A_c\x(t) + B_c\u(t)+\v(t), \quad \y(t) = C_c\x(t) + \w(t),
\end{equation}
where $\x(t) \in \reals^d$ is the state, $\u(t) \in \reals^\ell$ is the control input, $\y(t)\in \reals^{d_w}$ is the workspace output, and $\v \sim \mathcal{N}(\mathbf{0}, V_c)$ and $\w \sim \mathcal{N}(\mathbf{0}, W_c)$
represent Gaussian process and measurement noise, respectively. Fix a tracking controller timestep $\dt$ so that, e.g., $\x_t$ denotes $\x(t\cdot\dt)$, and fix a nominal trajectory $(\xn(t), \un(t), \yn(t))$, $t \in [0, \tau]$ satisfying the dynamics \eqref{eq:cont} without the noise terms $\v$ and $\w$. With deviation variables defined as $\dx_t:=\x_t - \xn_t$, $\du_t:=\u_t - \un_t$, and $\dy_t:=\y_t - \yn_t$, for $t = 0,\dots,T$, the \revv{discretized approximate dynamics for the tracking controller evolve as} 
\begin{equation}
\label{eq:disc}
\small
\begin{aligned}
&\dx_{t+1} = A\dx_t + B\du_t+\v_t, \quad \v_t \sim \mathcal{N}(\mathbf{0},
V), \\
&\dy_t = C\dx_t + \w_t, \quad \w_t \sim \mathcal{N}(\mathbf{0}, W),\\
&A := e^{A_c\dt}, \ \ B := \left(\int_0^\dt e^{A_cs} ds\right)B_c, \ \ C := C_c, \\
&V := \int_0^\dt e^{A_cs}V_c e^{A^\top_cs} ds, \ \ W := \frac{W_c}{\dt}.
\end{aligned}
\end{equation}
The discrete LQG controller $\du^{\text{LQG}}_t := L_t \, \widehat \dx_t$, with $L_t$ and $\widehat \dx_t$ denoting the feedback gain matrix and Kalman state estimate respectively (see \cite{ShaijuPetersen2008} for computation details), minimizes the tracking cost
$
J := \mathbb{E}\left[\dx_T^\top F \dx_T + \sum_{t=0}^{T-1} \dx_t^\top Q \dx_t + \du_t^\top R \du_t \right].
$
Then the control in continuous time is $\u(t) = \un(t) + \du^{\text{LQG}}_{\lfloor t/\dt\rfloor}$.

The SKP problem is posed as a cost optimization over dynamically-feasible trajectories (also referred to as motion plans) subject to a safety tolerance constraint separate from the optimization objective. Let $\xobs$ be the obstacle space, so that $\xfree = \mathbb{R}^d\setminus\xobs$ is the free state space. Let $\xgoal \subset \xfree$ and $\x_0 = \xinit \in \xfree$ be the goal region and initial state, respectively. Given a trajectory cost measure $c$ and CP tolerance $\alpha$, we wish to solve the following \skp.\\

\noindent\textbf{Stochastic Kinodynamic Motion Planning (\skp):}
\begin{equation}
\label{eq:skmpp}
\small
\begin{split}
\min_{\un(\cdot)} & \quad c(\xn(\cdot)) \\
\text{s.t.} & \quad \prob{\{x(t) \mid t \in [0,\tau]\} \cap \xobs \neq \varnothing}  \le \alpha\\
& \quad \u(t) = \un(t) + \du^{\text{LQG}}_{\lfloor t/\dt\rfloor}\\
& \quad \text{Equation \eqref{eq:cont}}.
\end{split}
\end{equation}
The numerical experiments in this work take $c$ as the mixed time/quadratic control effort penalty studied in \cite{WebbBerg2013,SchmerlingJansonEtAl2015b}. We make an implicit assumption in optimizing over the cost of the reference trajectory $c(\xn(\cdot))$ that the tracking control costs are minor in comparison to the nominal control cost; as noted in Section~\ref{sec:intro} this assumption also underlies the use of a tracking control strategy rather than computing a full policy over state beliefs.

\section{Collision Probability Approximation}\label{sec:cpapprox}

This section discusses a key challenge in computing solutions for the SKP problem, that of efficiently and accurately assessing collision probability. As discussed in \cite{JansonSchmerlingEtAl2015b}, Monte Carlo methods exist as an asymptotically \emph{exact} means for calculating the CP when tracking a trajectory segment. These methods are suitable for validating the most promising candidate trajectories post-exploration, but the high branching factor of multiobjective search necessitates less computationally intensive approximation schemes when checking a great number of candidate motions (most of which are discarded as Pareto dominated) in a tempo compatible with real-time operations.
In Section~\ref{sec:addmultcp} we review existing strategies for approximating CP. In Section~\ref{sec:hsmc} we present the novel Half-Space Monte Carlo (\hsmc) method aimed at efficiently and accurately approximating CP, particularly on GPUs, and analyze this method with numerical experiments. 

\subsection{\rev{Additive, Multiplicative, and MC CP Approximations}}\label{sec:addmultcp}

While the constraint in \eqref{eq:skmpp} is defined over continuous trajectory realizations, there are a number of waypoint-based schemes which derive an approximation of trajectory CP as a function of the instantaneous pointwise collision probabilities at the intermediate waypoints $\{\x_t\}$. \emph{Additive} approximations \cite{LudersKothariEtAl2010}
employ a conservative union bound
\begin{equation}\label{eq:addapprox}
\small
\text{CP} \approx \prob{\bigvee_{t = 0}^T \{\x_t \in \xobs\}} \leq \sum_{t = 0}^T \prob{\{\x_t \in \xobs\}},
\end{equation}
while \emph{multiplicative} approximations \cite{BergAbbeelEtAl2011}
assume independence between pointwise collision events: 
\begin{equation}
\small
1 - \text{CP} \approx \prob{\bigwedge_{t = 0}^T \{\x_t \notin \xobs\}} \approx \prod_{t = 0}^T \prob{\{\x_t \notin \xobs\}}.
\end{equation}
This independence assumption may be relaxed by fitting Gaussians successively to the waypoint distributions conditioned on no prior collisions (see, e.g., \cite{SunPatilEtAl2015,PatilBergEtAl2012} for details), in order to approximate the terms of the product expansion
\begin{equation}
\small
\begin{aligned}
1 - \text{CP} &\approx \prob{\bigwedge_{t = 0}^T \{\x_t \notin \xobs\}} \\
              &= \prod_{t = 0}^T \probcond{\{\x_t \notin \xobs\}}{\bigwedge_{s = 0}^{t-1} \{\x_s \notin \xobs\}},
\end{aligned}
\end{equation}
which we refer to in this text as the \emph{conditional multiplicative} approximation.
A common additional approximation made with these waypoint-based approaches is to assume a simplified collision model, for example by approximating the free space around a waypoint as an ellipsoid \cite{LudersKothariEtAl2010} or by a locally convex region defined by half-space constraints \cite{PatilBergEtAl2012}. We note that even though these collision models, as well as the additive approximation~\eqref{eq:addapprox}, may be deemed conservative, none of these methods are guaranteed to over- or under-approximate the actual CP as they do not allow for collisions in the intervals between waypoints \cite{JansonSchmerlingEtAl2015b}.

Monte Carlo methods can exactly address the dependencies between waypoint CPs, as well as accommodate continuous collision checking, at the expense of computational effort and statistical variance in the estimate. In \cite{JansonSchmerlingEtAl2015b}, variance-reduced MC methods are developed for estimating CP along the span of a full trajectory. The method in \cite{Agha-mohammadiChakravortyEtAl2014} breaks dependence between collision probabilities at certain points in the state space using feedback controllers. This construct divides the CP computation into smaller pieces corresponding to local connections, which allows for expensive MC-based CP computation to be done offline, given the environment map.

\subsection{Half-Space Monte Carlo}\label{sec:hsmc}

\begin{figure}[h]
    \centering
    \hspace*{\fill}%
    \begin{subfigure}[b]{0.2\textwidth}
        \includegraphics[width=\textwidth]{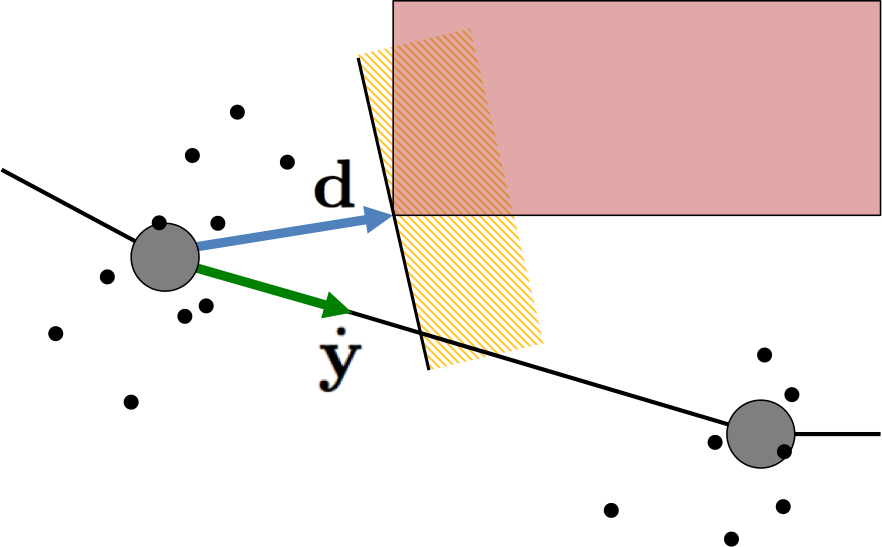}
        \caption{}
        \label{fig:hsformationVanilla}
    \end{subfigure}\hfill%
    \begin{subfigure}[b]{0.2\textwidth}
        \includegraphics[width=\textwidth]{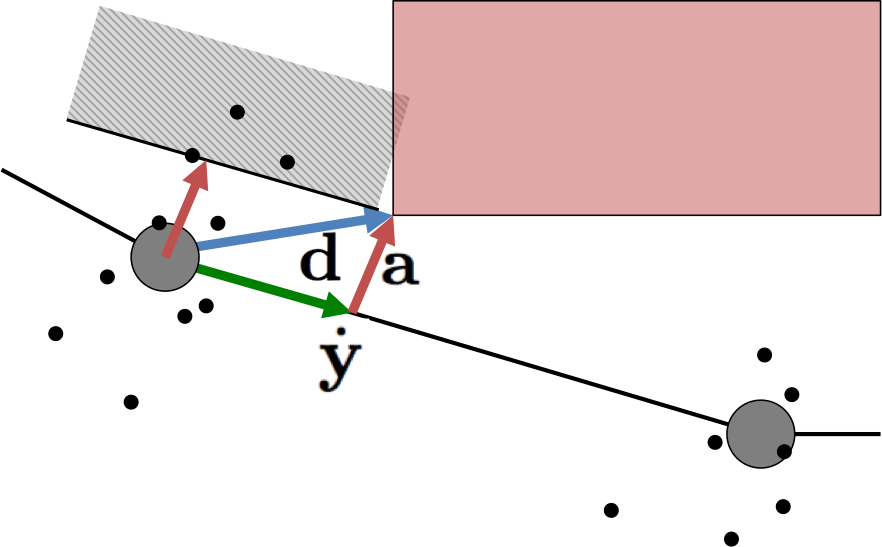}
        \caption{}
        \label{fig:hsformationAlt}
    \end{subfigure}     
    \hspace*{\fill}%   
    \caption{The vector $\d$ (blue) is the vector of minimum distance connecting the waypoint to the obstacle set and $\dot{\y}$ (green) is the projected velocity onto the workspace. 
    (\ref{fig:hsformationVanilla}) Illustration of half-space formation by distance only. 
    % (\ref{fig:hsformationAlt}) Illustration of the alteration of the half-space to account for motion; the vector $\a$ (red) is computed as ${\a} = {\d} - (\frac{{\d} \cdot {\dot{\y}}}{{\dot{\y}} \cdot {\dot{\y}}}) {\dot{\y}}$.
    (\ref{fig:hsformationAlt}) Illustration of the alteration of the half-space \rev{that accounts for the notion that a particle is in collision with an obstacle only if it is both close and its motion is towards the obstacle}; the vector $\a$ (red) is computed as ${\a} = {\d} - (\frac{{\d} \cdot {\dot{\y}}}{{\dot{\y}} \cdot {\dot{\y}}}) {\dot{\y}}$.}
\label{fig:hsformation}
\end{figure}

In this subsection we detail the massively parallel \hsmc method and show that it attains good empirical accuracy. 
This CP approximation is inspired by particle-based MC methods in their ability to represent arbitrary state distributions and how they evolve in $\xfree$, as well as by the efficient collision models of waypoint-based methods. 
Briefly, \hsmc is a Monte Carlo method where, within local convex regions around a reference waypoint $\xn_t$, sampled trajectories (particles) are checked at each time step by removing those that might be ``expected'' to be in collision through subsequent motion towards $\xn_{t+1}$. 
Instead of computing a full collision check, we only compare the particle's position in the {\em workspace} against the half-space boundaries of the local convex region 
(\rev{with consideration for the reference waypoint state's velocity, see Fig.~\ref{fig:hsformation}).} 
The local convexification process employed by \hsmc differs from that presented in \cite{PatilBergEtAl2012} and used in \cite{JansonSchmerlingEtAl2015b}, in that obstacle half-spaces are computed on the basis of Euclidean distance in the workspace, as opposed to Mahalanobis distance. 
In those works Mahalanobis distance is used as a proxy for the shape of the uncertainty distribution centered at $\xn_t$; \hsmc instead maintains a discretized Monte Carlo representation of this uncertainty distribution so that position and velocity are sufficient to judge approximate collision in the Euclidean workspace. 
We remark that the computation of each local convex region is independent of trajectory history and uncertainty distribution, thus allowing for a massively parallel computation at the start of planning. 
Furthermore, if the obstacles are available a priori, this computation can be performed entirely offline.

As we are restricting our attention to the workspace, recall that $\y_t = C \x_t$ is the projection of the state into the workspace.
The construction of the local convex region around $\yn_t$ is performed using a sequential process, following the work of \cite{PatilBergEtAl2012}. 
At step $i$, we define $\d_{t,i}$ as the vector from $\yn_t$ to the closest obstacle point and prune away \rev{all obstacle geometric structures} that lie in the half-space $\d_{t,i}^\top (\z - \yn_t) > \d_{t,i} \cdot \d_{t,i}$, where $\z \in \reals^{d_w}$. 
This procedure continues until all obstacle geometry has been \rev{pruned away}. 
The process defines the local convex region, however the final approximate collision check should represent the notion that a particle will collide with an obstacle only if it is both close and its motion is towards the obstacle.
We thus alter these half-spaces, as illustrated in Fig.~\ref{fig:hsformation}, by projecting them to be perpendicular to the direction of travel via
\begin{equation}
\small
{\a_{t,i}} = {\d_{t,i}} - \bigg(\frac{{\d_{t,i}} \cdot {\dot{\y}_t}}{{\dot{\y}_t} \cdot {\dot{\y}_t}}\bigg) {\dot{\y}_t}.
\label{eq:hs}
\end{equation}
The resulting collision constraint for an MC particle is then defined by $\a_{t,i}^\top \dy_t > \b_{t,i}$, where $\b_{t,i} = \a_{t,i} \cdot \a_{t,i}$.
That is, to approximate the CP of a given motion plan we generate a set of $N$ particle trajectories and compare their deviations $\dy_t$ from the nominal trajectory at each time step to the local collision half-spaces. 
Every half-space check $(\a_{t,i}, \b_{t,i})$ along a particle trajectory is independent of every other, requiring only a final OR reduction to determine the validity of a particle, thus this process is amenable to parallelization. The final CP approximation is obtained by counting the number of invalid particle trajectories and dividing by $N$.

\begin{figure}[b]
    \centering
    \begin{subfigure}[b]{0.22\textwidth}
        \includegraphics[width=\textwidth]{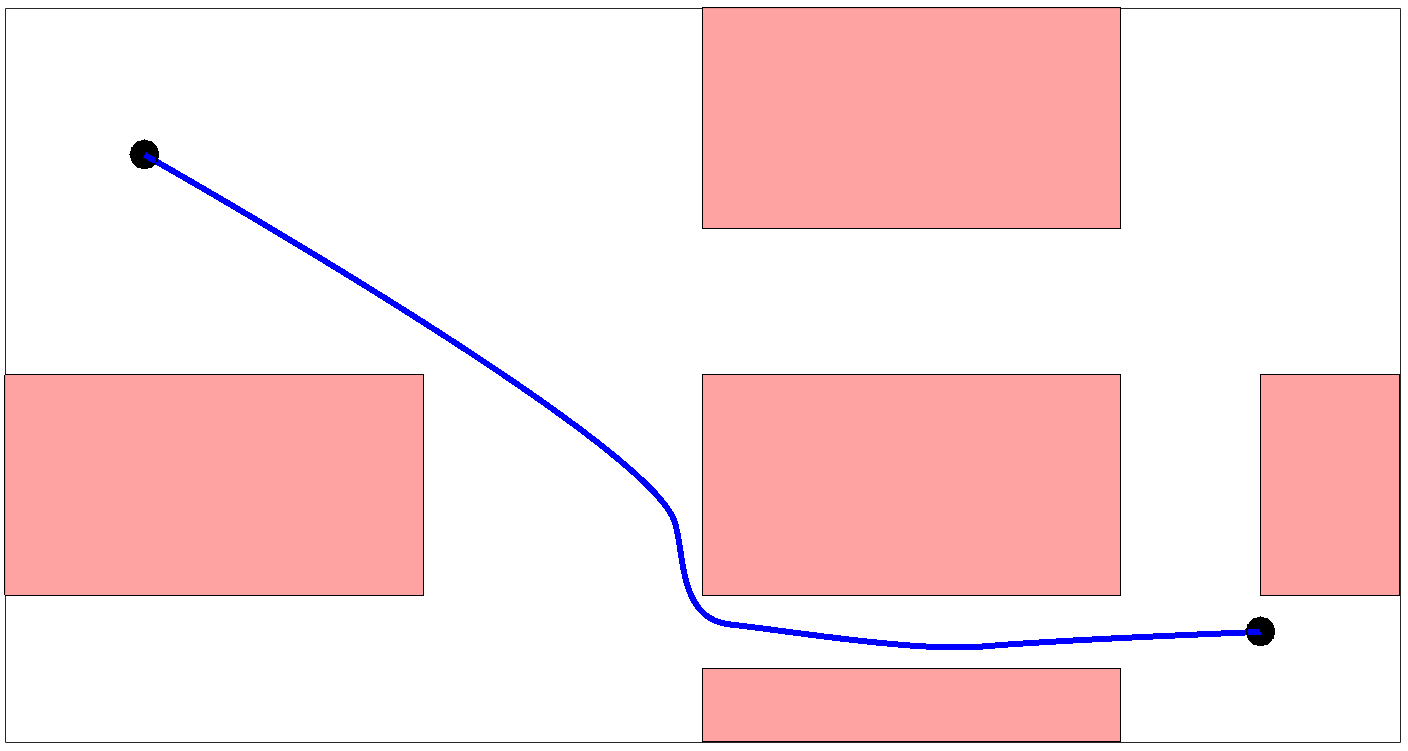}
        \caption{}
        \label{fig:cpCompare_p04t05s}
    \end{subfigure}
    \begin{subfigure}[b]{0.25\textwidth}
        \includegraphics[width=\textwidth]{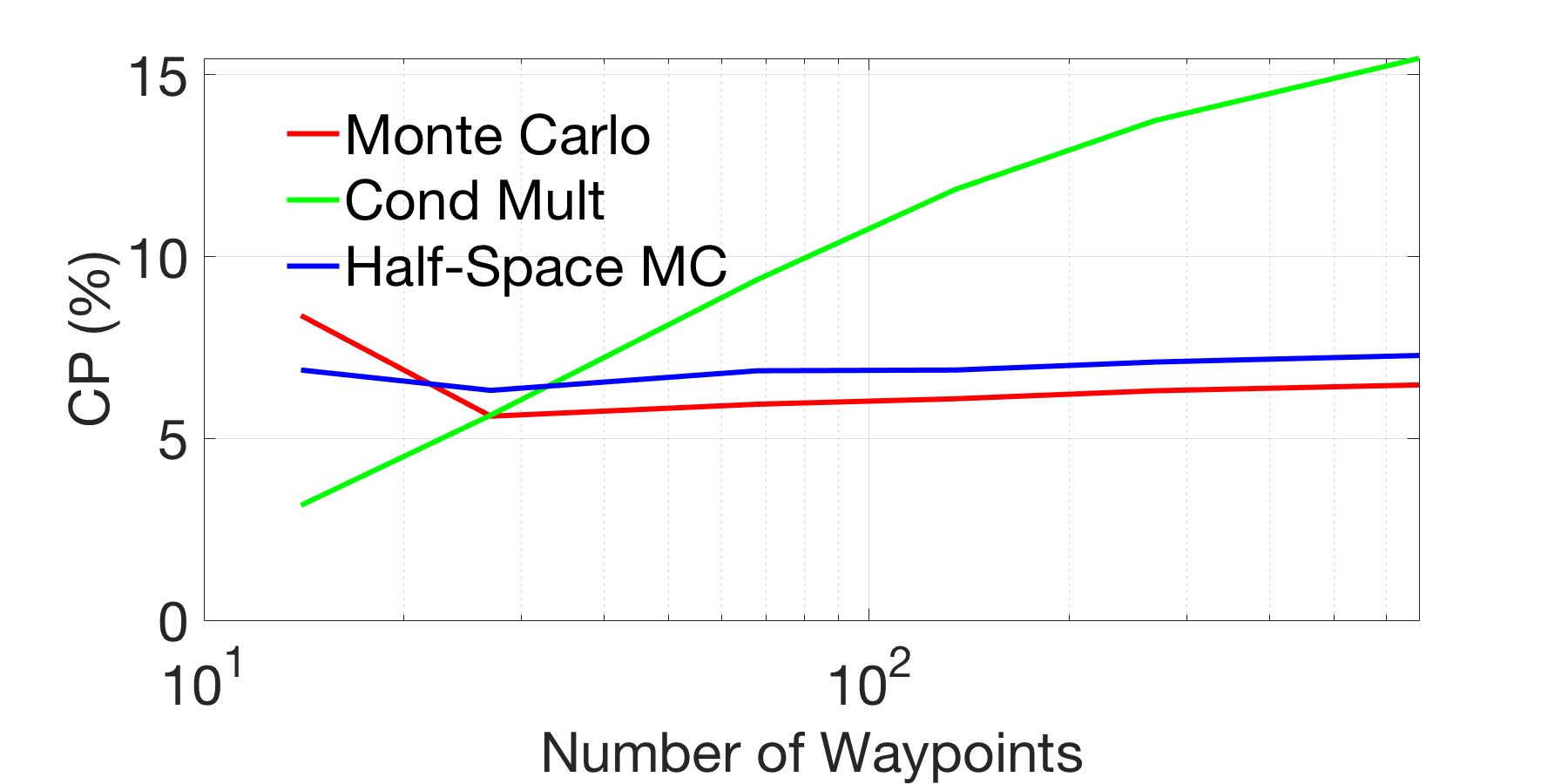}
        \caption{}
        \label{fig:cpCompare_p04t05}
    \end{subfigure}
    \begin{subfigure}[b]{0.22\textwidth}
        \includegraphics[width=\textwidth]{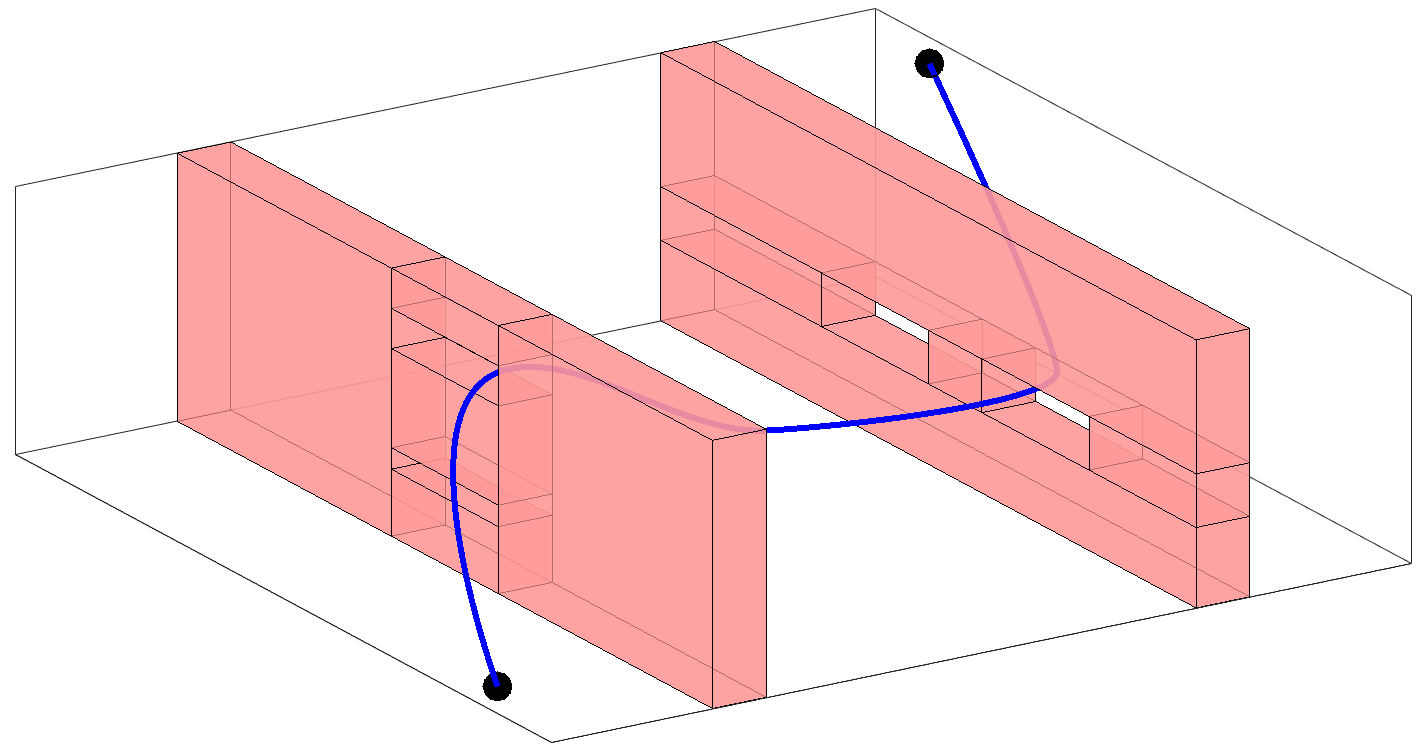}
        \caption{}
        \label{fig:cpCompare_p05t01s}
    \end{subfigure}    
    \begin{subfigure}[b]{0.25\textwidth}
        \includegraphics[width=\textwidth]{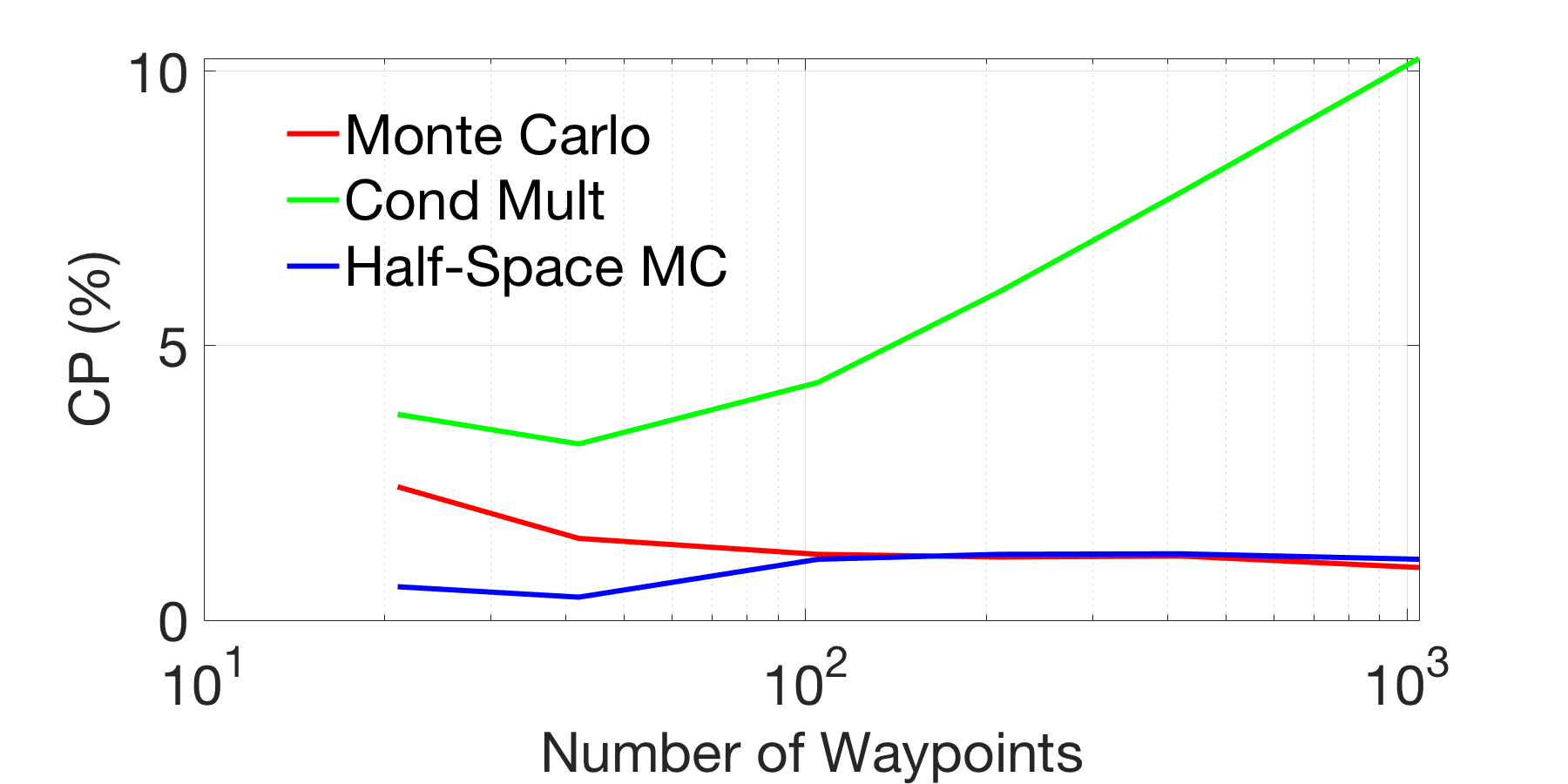}
        \caption{}
        \label{fig:cpCompare_p05t01}
    \end{subfigure}
    % \begin{subfigure}[b]{0.23\textwidth}
    %     \includegraphics[width=\textwidth]{cpCompare_p05t20cs}
    %     \caption{}
    %     \label{fig:cpCompare_p05t20cs}
    % \end{subfigure}
    % \begin{subfigure}[b]{0.23\textwidth}
    %     \includegraphics[width=\textwidth]{cpCompare_p05t20c}
    %     \caption{}
    %     \label{fig:cpCompare_p05t20c}
    % \end{subfigure}
    % \begin{subfigure}[b]{0.23\textwidth}
    %     \includegraphics[width=\textwidth]{cpCompare_p06t10s}
    %     \caption{}
    %     \label{fig:cpCompare_p06t10s}
    % \end{subfigure}
    % \begin{subfigure}[b]{0.23\textwidth}
    %     \includegraphics[width=\textwidth]{cpCompare_p06t10}
    %     \caption{}
    %     \label{fig:cpCompare_p06t10}
    % \end{subfigure}
\caption{Comparison of the true CP (Monte Carlo, red), conditional multiplicative CP approximation (green), and \hsmc CP approximation (blue) versus number of waypoints (i.e., as the timestep $\dt$ decreases) over several problem instances. Each \hsmc approximation and MC CP was computed with 10,000 samples. Note that the conditional multiplicative approximation tends to become increasingly conservative as the number of waypoints increases, whereas \hsmc generally remains near the true CP.}
\label{fig:cpCompare}
\end{figure}

The results in Fig.~\ref{fig:cpCompare} show that with two representative planning problems, targeting CPs of 1\% and 5\%, \hsmc outperforms conditional multiplicative in terms of approximation accuracy. 
While the \hsmc CP approximation improves as the number of waypoints increases, conditional multiplicative's conservative nature causes the CP approximation to degrade, in agreement with the results in \cite{JansonSchmerlingEtAl2015b}.
\vspace{0.3cm}

\section{Parallel Uncertainty-aware \\Multiobjective Planning}\label{sec:pump}

The Parallel Uncertainty-aware Multiobjective Planning (\pump) algorithm explores the state space via a multiobjective search, building a Pareto set of motion plans from the initial state to goal region, where Pareto dominance is determined by cost and approximated CP. 
This Pareto set comprises not only several motion plans, but different solution homotopies, which are searched to find the minimum cost plan certified to satisfy a CP constraint. 
The algorithm is briefly outlined in Alg.~\ref{alg:PUMPoutline} and formally stated in Algs.~\ref{alg:PUMP}-\ref{alg:PUMPsel}.

\begin{algorithm}
\small
\caption{\pump: Outline}
\label{alg:PUMPoutline}
\algsetup{linenodelimiter=}
\begin{algorithmic}[1]
\STATE Massively parallel sampling-based graph building
\STATE Multiobjective search to find $\Psoln :=$ Pareto optimal plans in terms of (cost, approximate CP)
\STATE Bisection search $\Psoln$ to find the best motion plan, $p^*$, with a verified CP below constraint\\
$\quad$ If feasible plan found, locally optimize (smooth) \\ 
$\quad$ subject to CP constraint satisfaction and return $p^*$\\
$\quad$ Otherwise, report failure
\end{algorithmic}
\end{algorithm}

To aid in formal description, we now present some notation and algorithmic primitives.
In line with the sampling-based motion planning literature, we construct motion plans as a sequence of sampled nodes (samples) in $\mathbb{R}^d$.
Since we are interested in tracking collision probability, which is history dependent, we store plans as structs with attributes $(\mathrm{head},\mathrm{path},\mathrm{cost},\hat{\mathrm{cp}})$. The fields represent the terminal trajectory node from which we may extend other plans, the list of previous nodes, the cost, and the approximate CP, respectively. 
Let $\texttt{SampleFree}(n)$ be a function that returns a set of $n \in \N$ points sampled from $\xfree$. 
In the following definitions let $u,v \in \xfree$ be samples, and let $V \subset \xfree$ be a set of samples.
Define $\texttt{Cost}(u,v)$ as a function that returns the cost of the optimal trajectory from $u$ to $v$. 
Let $\texttt{Near}(V,u,r)$ be a function that returns the set of samples $\{v \in V : \texttt{Cost}(u,v) < r\}$. 
Let $\texttt{Collision}(u,v)$ denote whether the optimal trajectory from $u$ to $v$ intersects $\xobs$. 
Given a set of motion plans $P$ and $\Popen \subset P$, $\texttt{RemoveDominated}(P,\Popen)$ denotes a function that removes \rev{all} $p \in \Popen$ from $\Popen$ and $P$ that are dominated by \rev{a motion plan in $\{p_\text{dom} \in P : p_\text{dom}.\text{head} = p.\text{head}\}$; \rev{explicitly}, we say that $p_\text{dom}$ dominates $p$ if \revv{$(p.\text{cost} > p_\text{dom}.\text{cost}) \wedge (p.\hat{\text{cp}} \geq p_\text{dom}.\hat{\text{cp}})$}}.
Let $\widehat{\texttt{CP}}(v,p)$ denote a function that returns an approximate CP for the motion plan $p$ concatenated with the optimal connection $p.\text{head}$ to $v$. We define $\texttt{MC}(p)$ as a function that returns an asymptotically exact CP estimate of motion plan $p$ through Monte Carlo sampling \cite{JansonSchmerlingEtAl2015b}. 
Since \pump conducts its exploration within the context of a discrete graph representation of $\xfree$, the best costs and $\hat{\text{cp}}$s are limited by the resolution of the graph. 
To address variance in sample placement, we include a post-processing ``smoothing'' step which locally optimizes the final plan subject to CP constraint recertification.
 %(\rev{i.e., the constraint that the $\text{CP}$ of the smoothed plan is still  less than or equal to $\alpha$}). 
 We denote this process by $\texttt{Smooth}(p, \alpha)$; in this work we use the strategy in \cite{StarekSchmerlingEtAl2016b}, which performs a bisection search to create a new smoothed plan as a combination of the original plan and the optimal unconstrained trajectory from $\xinit$ to $p.\text{head}$.

We are now ready to fully detail the \pump algorithm (Algs.~\ref{alg:PUMP}-\ref{alg:PUMPsel}). 
The main body is provided in Alg.~\ref{alg:PUMP}; for brevity we assume the algorithm inputs and tuning parameters are available globally to the subroutines Algs.~\ref{alg:PUMPpre}-\ref{alg:PUMPsel}.
\pump can be divided into three phases. 
The first is a graph building phase (Alg.~\ref{alg:PUMPpre}), similar to probabilistic roadmap methods \cite{AmatoDale1999}, that constructs a representation of available motions within the free state space.
This phase begins by adding $\xinit$ and a set of $n$ samples from $\xfree$ (including at least one from $\xgoal$) to the set of samples $V$. 
Each sample's nearest neighbors are then computed and each neighbor edge is collision checked, storing the collision-free neighbors of $v$ in $N(v)$. 
\revv{Here one should consider the ``kinodynamic counterparts" of traditional Euclidean nearest neighbors and straight line edges \cite{SchmerlingJansonEtAl2015b,SchmerlingJansonEtAl2015}. Specifically, for the SKP setting of this paper, we refer to nearest neighbors as samples within a connection radius $r_n$, where $r_n$ is a tuning parameter, which we set as described in \cite{SchmerlingJansonEtAl2015b}.
Depending on the CP approximation strategy used during exploration, this graph building stage can also calculate obstacle half-spaces at each waypoint.
Following \cite{AmatoDale1999}, this phase is embarrassingly parallel.}

\begin{algorithm}[t]
\small
\caption{\pump}
\label{alg:PUMP}
\algsetup{linenodelimiter=}
\begin{algorithmic}[1]
\REQUIRE planning problem $(\xfree, \xinit, \xgoal)$, number of nodes $n \in \mathbb{N}$, connection radius $r_n \in \mathbb{R}_{\geq 0}$, CP threshold $\alpha \in (0,1)$
\ENSURE CP approx. factor $\eta > 1$, group cost factor $\dr \in (0, 1]$
\STATE $(V,N) = \texttt{BuildGraph}()$ \COMMENT{Alg.~\ref{alg:PUMPpre}}
\STATE $\alpha_\mathrm{min} = \frac{1}{\eta}\cptarget$; $\alpha_\mathrm{max} = \eta\cptarget$
\STATE $\Psoln \leftarrow \texttt{Explore}(\alpha_\mathrm{min}, \alpha_\mathrm{max}, V, N)$ \COMMENT{Alg.~\ref{alg:PUMPexp}}
\STATE $p^* \leftarrow \texttt{PlanSelection}(\alpha,\Psoln)$ \COMMENT{Alg.~\ref{alg:PUMPsel}}
\RETURN $p^*$
\end{algorithmic}
\end{algorithm}

\begin{algorithm}[t]
\small
\caption{\texttt{BuildGraph}}
\label{alg:PUMPpre}
\algsetup{linenodelimiter=}
\begin{algorithmic}[1]
\STATE $V \leftarrow \{\xinit\} \union \texttt{SampleFree}(n)$
\FORALL[massively parallel graph building]{$v \in V$}
\STATE $N(v) \leftarrow \texttt{Near}(V \setminus \{v\}, v, r_n)$
\FORALL{$u \in N(v)$}
\rev{\LINEIF{$\texttt{Collision}(v,u)$}{$N(v) \leftarrow N(v) \setminus \{u\}$}}
\ENDFOR
\ENDFOR
\RETURN ($V$, $N$)
\end{algorithmic}
\end{algorithm}

\begin{algorithm}[t]
\small
\caption{\texttt{Explore}$(\alpha_\mathrm{min}, \alpha_\mathrm{max}, V, N)$}
\label{alg:PUMPexp}
\algsetup{linenodelimiter=}
\begin{algorithmic}[1]
\STATE $\Popen \leftarrow \{(\xinit, \emptyset, 0, 0)\}$    \hspace{0.cm} // plans ready to be expanded
\STATE $P(\xinit) \leftarrow \Popen$                        \hspace{0.99cm} // plans with head at $\xinit$
\STATE $\B \leftarrow \Popen$                               \hspace{1.88cm} // plans considered for expansion
\STATE $i = 0$
\WHILE{$\Popen \neq \emptyset \land \{g \in \B : (g.\mathrm{head} \in \xgoal) \wedge $\\$\qquad
    (g.\hat{\mathrm{cp}} < \alpha_\mathrm{min})\} = \emptyset$}\label{line:exploreWhile}
\FORALL{$p \in \B$}
\FORALL{$x \in N(p.\mathrm{head})$}
\STATE $q \leftarrow (x, p.\mathrm{path} + \{p.\mathrm{head}\}, p.\mathrm{cost} + $\\$\quad \texttt{Cost}(p.\mathrm{head},x),\widehat{\texttt{CP}}(x,p))$\label{line:newPath}
\IF[CP cutoff]{$q.\hat{\mathrm{cp}} < \alpha_\mathrm{max}$}
\STATE $P(x) \leftarrow P(x) \union \{q\}$
\STATE $\Popen \leftarrow \Popen \union \{q\}$
\ENDIF
\ENDFOR
\ENDFOR
\STATE $(P,\Popen) \leftarrow \texttt{RemoveDominated}(P,\Popen)$
\STATE $\Popen \leftarrow \Popen \setminus \B$
\STATE $i \leftarrow i + 1$
\STATE $\B \leftarrow \{p \in \Popen : p.\mathrm{cost} \leq i \dr r_n\}$
\ENDWHILE
\RETURN $\Psoln \leftarrow \{p \in P(v) : v \in \xgoal\}$
\end{algorithmic}
\end{algorithm}

\begin{algorithm}[t]
\small
\caption{\texttt{PlanSelection}$(\alpha, \Psoln)$}
\label{alg:PUMPsel}
\algsetup{linenodelimiter=}
\begin{algorithmic}[1]
\STATE $P \leftarrow \texttt{Sort}(\Psoln)$     \hspace{0.5cm} // Sort $\Psoln$ in ascending $\widehat{\text{CP}}$ order
\STATE $l = 1, u = |\Psoln|, m = \lceil (l + u)/2 \rceil$
\WHILE{$l \neq u$}
\rev{\LINEIFELSE{$\texttt{MC}(P_m) > \alpha$}{$u = m - 1$}{$l = m$}}
\STATE $m = \lceil (l + u)/2 \rceil$
\ENDWHILE
\IF{$\texttt{MC}(P_m) > \alpha$}
\RETURN Failure                                 \hspace{0.6cm} // No feasible plans in $\Psoln$
\ENDIF
\RETURN $\texttt{Smooth}(P_m, \alpha)$
\end{algorithmic}
\end{algorithm}

The second phase, Alg.~\ref{alg:PUMPexp}, explores the state space to find the Pareto optimal set of motion plans, where Pareto dominance is determined in terms of cost and $\hat{\text{cp}}$. 
This phase takes an upper and lower bound, $\alpha_\mathrm{max}$ and $\alpha_\mathrm{min}$, expressed in terms of $\hat{\text{cp}}$.
These bounds are chosen as a multiplicative factor $\eta$ around the target CP to accommodate any under- or over-CP approximation; $\eta$ should be tuned according to the accuracy of $\widehat{\texttt{CP}}$, \revv{as discussed in detail in Section~\ref{sec:implementation}.}
\revv{Note too that \pump is left agnostic of CP approximation methodology, and while we use \hsmc for the numerical experiments in this work, its application is limited to workspace obstacles; if this is not the case for a given planning problem, other methods can be used \cite{BergAbbeelEtAl2011,PatilBergEtAl2012}.}
The exploration of the state space proceeds in a parallel fashion by expanding the group, $\B$, of all open (unexpanded) motion plans, $\Popen$, below an increasing cost threshold, to their nearest neighbors. 
During this process, any motion plans that are dominated or have $\hat{\text{cp}} > \alpha_\mathrm{max}$ are discarded.
The cost threshold is increased by $\dr r_n$ at each exploration loop, \revv{thus determining the size of $\B$ through the group cost factor $\dr \in (0,1]$; the choice of this value is discussed in Section~\ref{sec:implementation}}.
The exploration terminates once a plan reaches $\xgoal$ with $\hat{\text{cp}} < \alpha_\mathrm{min}$ or once $\Popen = \emptyset$, at which point all plans connecting $\xinit$ to $\xgoal$ are stored in $\Psoln$.
The $\alpha_\mathrm{min}$ early termination criterion introduces a tradeoff for $\dr$: choosing a high $\lambda$ increases algorithm speed through greater parallelism, but has the side effect of allowing $\Popen$ to spread out in cost, potentially terminating before a high quality partial plan has reached $\xgoal$.
Finally, we note that the discretized expansion groups has an added benefit of not requiring a heap to store a cost ordering on $\Popen$, but rather a \rev{bucket array} should be implemented, allowing $\mathcal{O}(1)$ insertion and removal.

Computational tractability of \pump relies on the use of the $\texttt{RemoveDominated}$ function, and thus it is important to discuss its associated assumptions. Pruning the search requires judging the potential of multiple candidate plans that arrive at the same node. We use each plan's $\hat{\text{cp}}$ for this purpose, but this disregards any impact of the shape of the plan's full probability distribution, conditioned on prior obstacle avoidance, at that node. 
Even an exact CP estimate would be insufficient for this purpose. 
The only remedy is an utterly exhaustive search; instead our goal during exploration is to produce a set of high-quality motion plans. 
We find that in practice, using $\hat{\text{cp}}$'s, this set encompasses all important trajectory homotopy classes. \footnote{We note here an additional potential source of algorithm suboptimality --- the fact that we search over a graph of only cost-optimal $u$-$v$ connections. In some cases, e.g., high initial uncertainty compared to the steady-state, it may be best to consider other strategies for local connection, e.g., waiting.}  % $10 says Marco removes this for being off-message, but I think it's extremely important

The final phase, Alg.~\ref{alg:PUMPsel}, performs a bisection search over $\Psoln$ to identify the lowest cost plan $p^*$ with a CP certifiably below $\alpha$, as verified through MC with full collision checking \cite{JansonSchmerlingEtAl2015b}.
If no plan is found, the algorithm reports failure.
Otherwise, $\texttt{Smooth}(p^*, \alpha)$ is returned, which reduces cost until any margin in the CP constraint is exhausted.

\section{Numerical Experiments}\label{sec:sims}

\subsection{Experimental Setup}\label{sec:implementation}

\rev{The simulations in this section use 6D double integrator dynamics ($\ddot{x} = u$) to model a quadrotor system in a 3D workspace, and were implemented in CUDA C and generated on an NVIDIA GeForce GTX 980 GPU on a Unix system with a 3.0 GHz CPU.}
Example code may be found at \url{github.com/StanfordASL/PUMP}.
\rev{We implemented state space sampling with the deterministic, low-dispersion Halton sequence, as motivated by the analysis in \cite{JansonIchterEtAl2015b}. Note that as this sampling is deterministic rather than random, variance is not reported.}
By preallocating GPU memory, and \rev{offline} precomputing and caching both nearest-neighbors and edge controls (which are independent of the particular \rev{obstacles set} $\xobs$), the total run time was significantly reduced.
\revv{The simulations below use the \hsmc CP approximation method detailed in Section~\ref{sec:hsmc} with $N=128$, noting that the number of particles may be quite low as it is solely used to guide the exploration phase towards promising motion plans rather than certify solutions.

Our algorithm introduces two tuning parameters, $\eta$ and $\dr$. 
The value of $\eta$, the CP approximation factor, represents a tradeoff between considering fewer plans (and thus run time) and the potential for either removal of promising plans or early exploration termination.
In practice, however, we have observed only a small influence on run time, and it should thus be set to the limits of accuracy for the CP approximation strategy.
Specific to \hsmc, we set $\eta = 2$ for target CPs above 1\% and higher values below 1\% to account for the high deviation factor from using a particle representation for rare events (for example at 0.1\% we use $\eta = 10$).
Lastly, for the group cost factor $\dr$, we have found that $\dr = 0.5$ represents a good balance of parallelism, ease of implementation, and potential early termination described in Section~\ref{sec:pump}.}

\subsection{Comparison to Monte Carlo Motion Planning}

\begin{figure}[h]
    \centering
    \begin{subfigure}[b]{0.21\textwidth}
        \includegraphics[width=\textwidth]{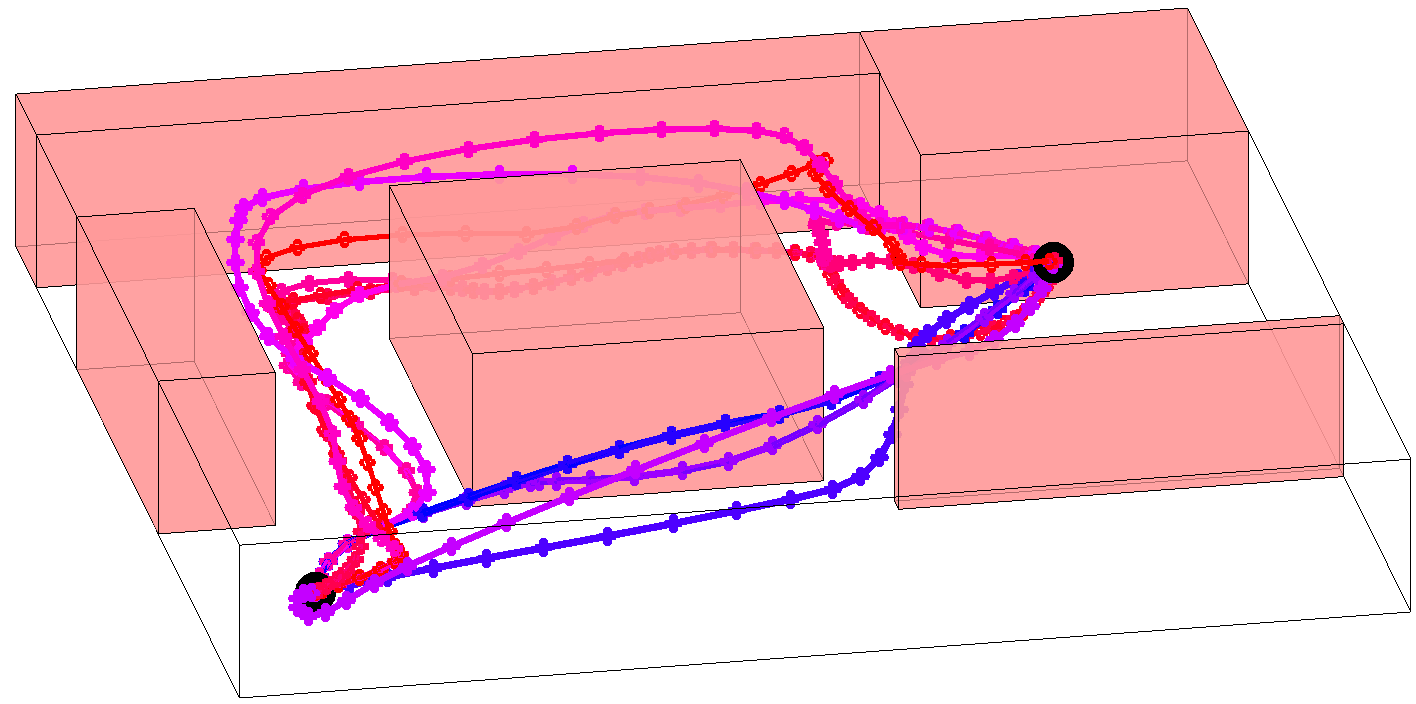}
        \caption{}
        \label{fig:mcmp_degradeSetup}
    \end{subfigure}    
    \begin{subfigure}[b]{0.21\textwidth}
        \includegraphics[width=\textwidth]{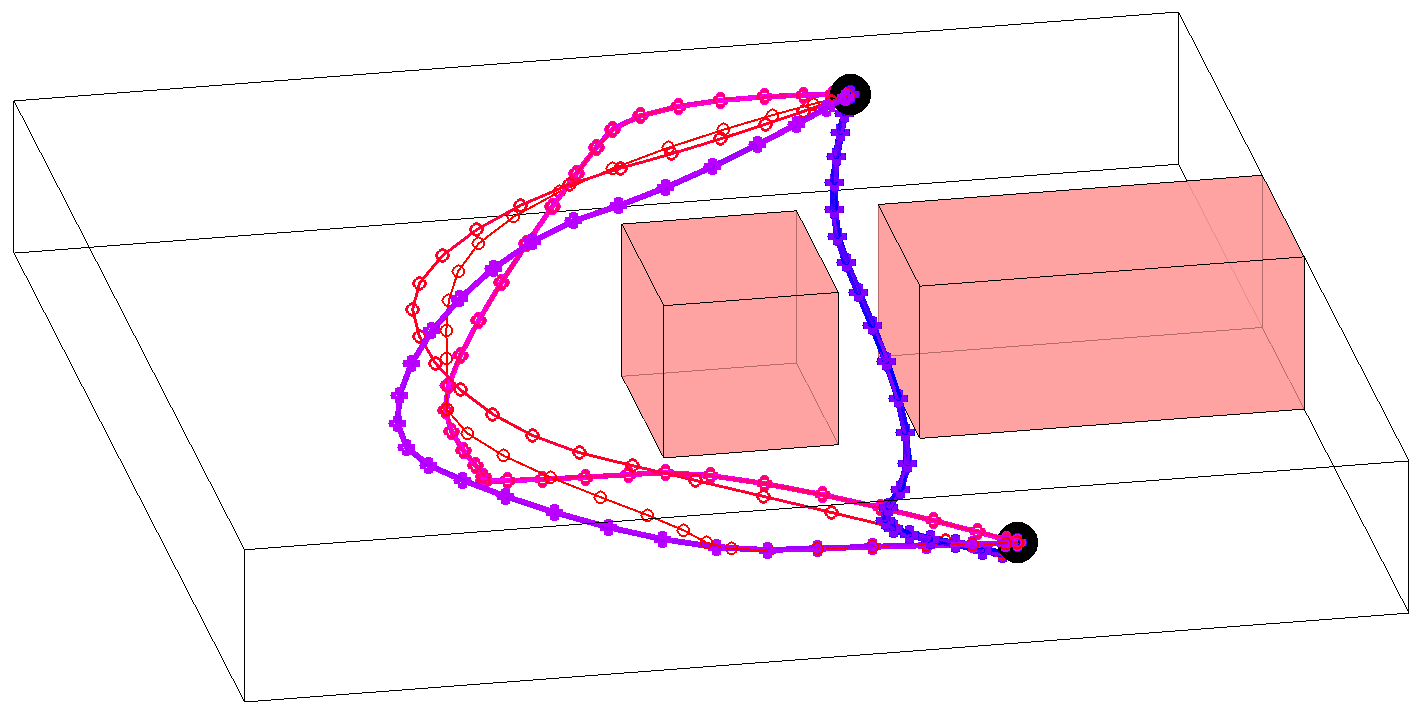}
        \caption{}
        \label{fig:mcmp_gapSetup}
    \end{subfigure}    
    \begin{subfigure}[b]{0.23\textwidth}
        \includegraphics[width=\textwidth]{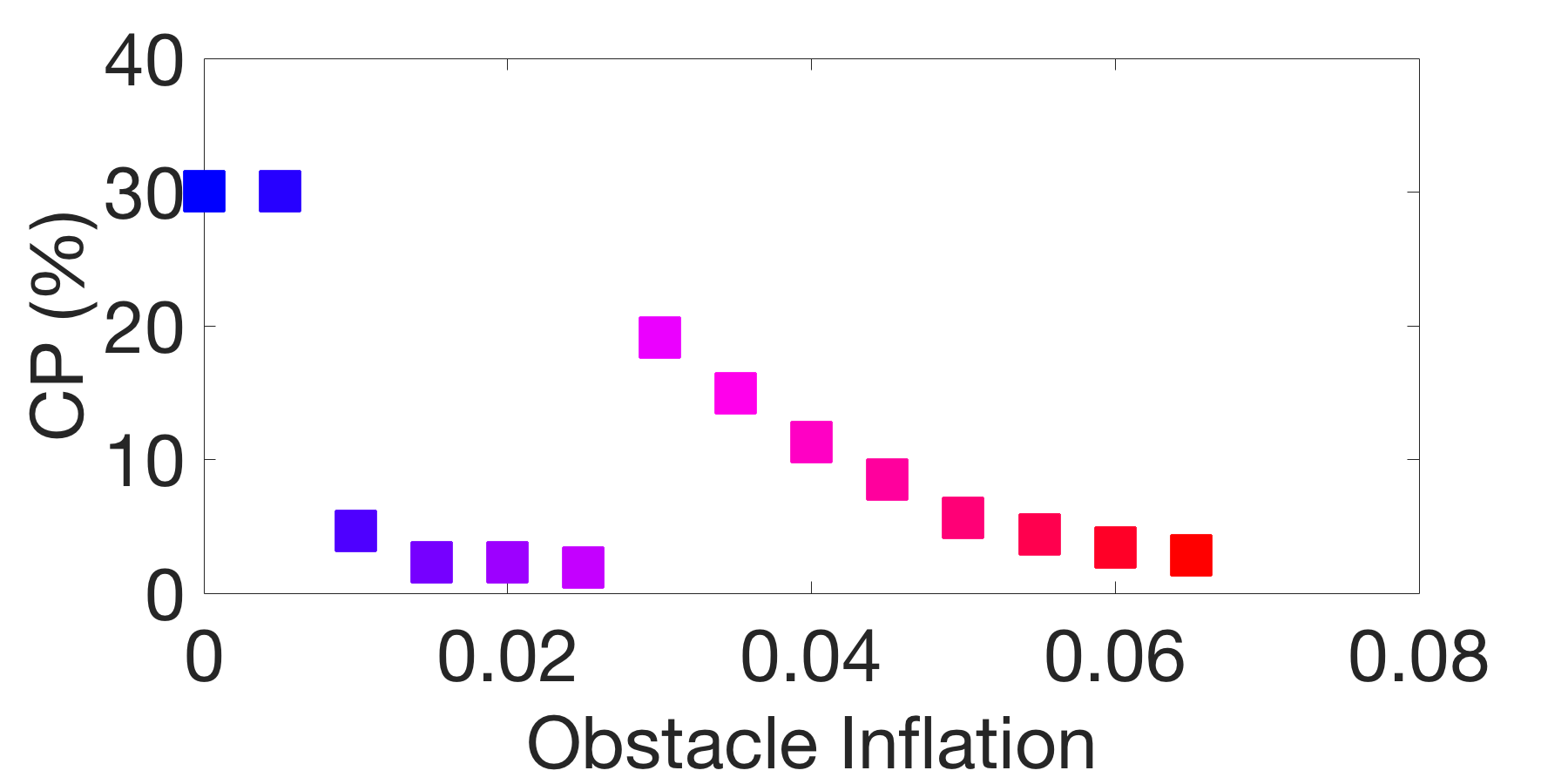}
        \caption{}
        \label{fig:mcmp_degradeInflation}
    \end{subfigure}
    \begin{subfigure}[b]{0.23\textwidth}
        \includegraphics[width=\textwidth]{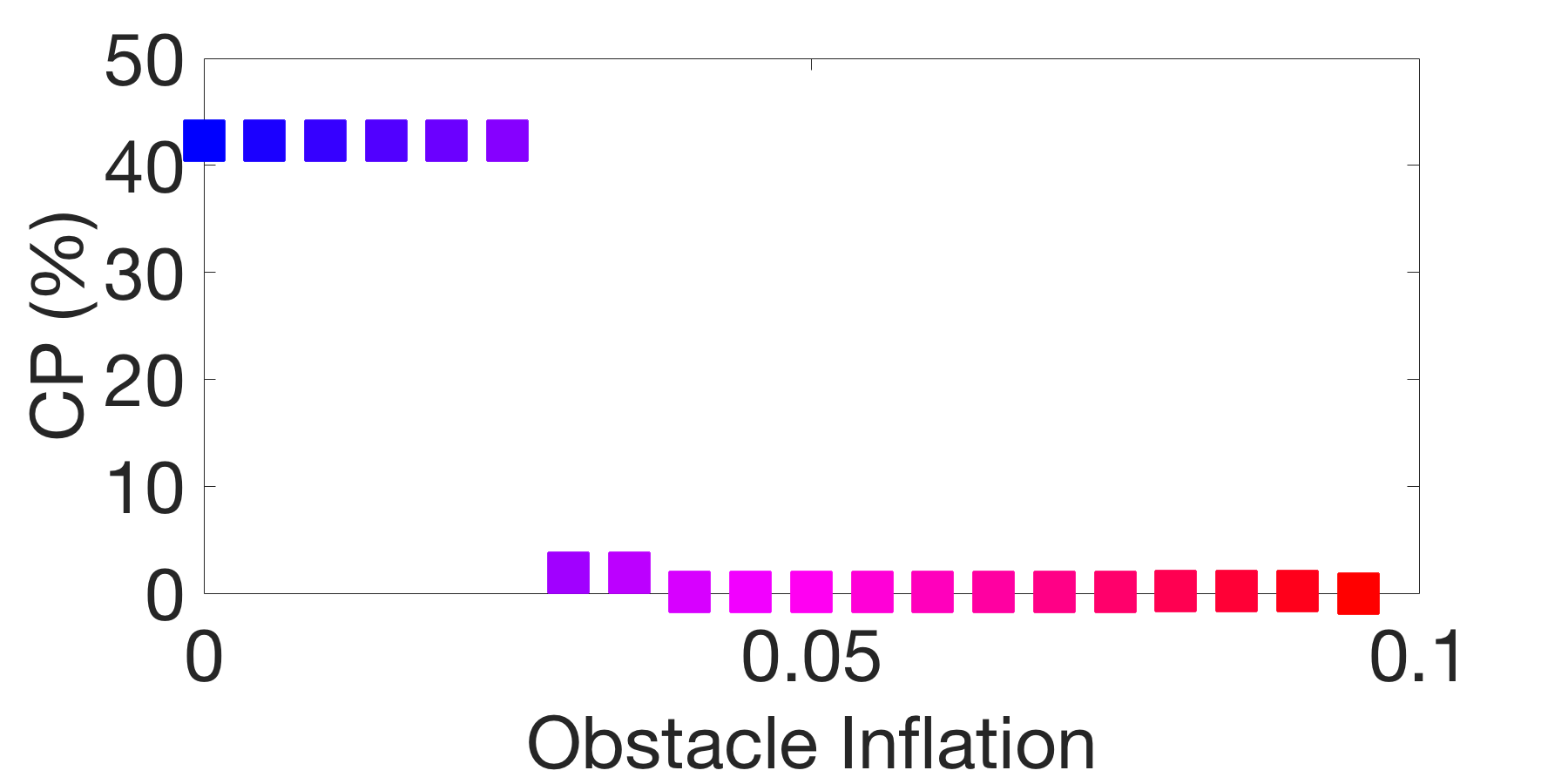}
        \caption{}
        \label{fig:mcmp_gapInflation}
    \end{subfigure}    
    \begin{subfigure}[b]{0.23\textwidth}
        \includegraphics[width=\textwidth]{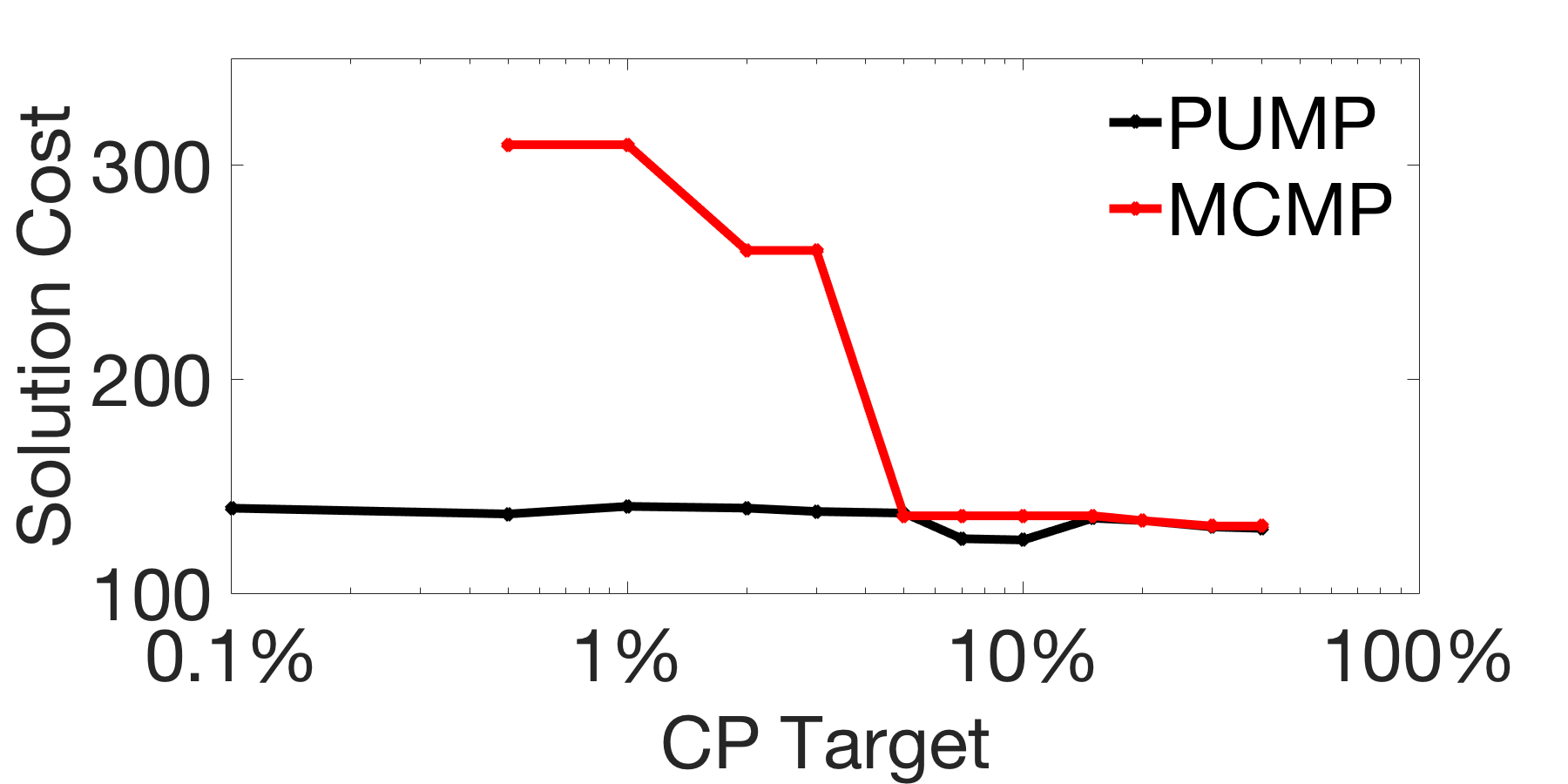}
        \caption{}
        \label{fig:mcmp_degradeCost}
    \end{subfigure}
    \begin{subfigure}[b]{0.23\textwidth}
        \includegraphics[width=\textwidth]{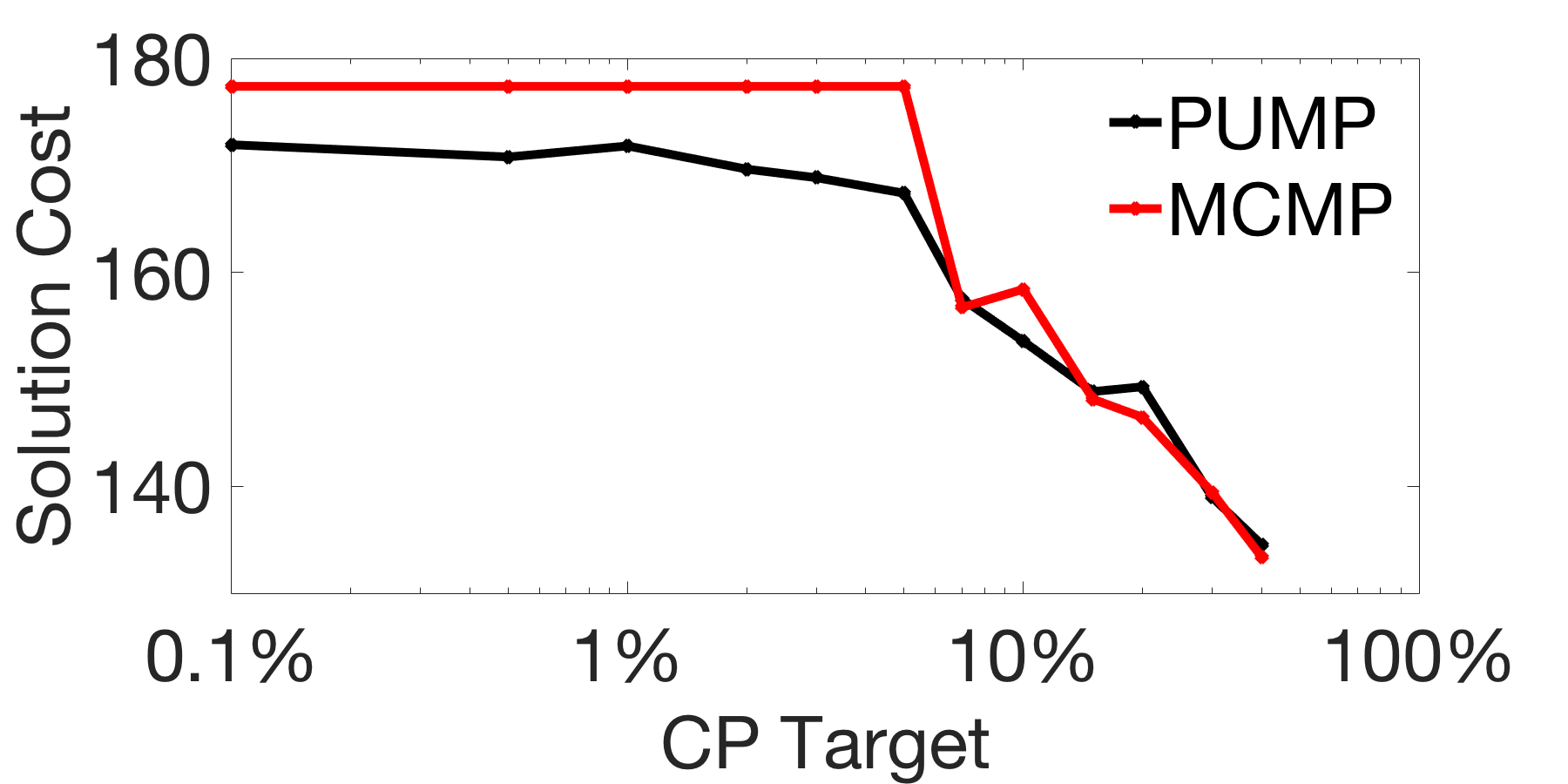}
        \caption{}
        \label{fig:mcmp_gapCost}
    \end{subfigure}    
\caption{\mcmp and \pump comparisons. (\ref{fig:mcmp_degradeSetup}, \ref{fig:mcmp_gapSetup}) plot MCMP candidate plans at a range of safety buffers. (\ref{fig:mcmp_degradeInflation}, \ref{fig:mcmp_gapInflation}) plot the CP of the returned solution \rev{to the deterministic problem with a given inflation factor.}
(\ref{fig:mcmp_degradeSetup}, \ref{fig:mcmp_degradeInflation}, \ref{fig:mcmp_degradeCost}) \revv{In this case CP has an increase as the inflation factor increases (via a discontinuous jump when the narrow gap closes at an inflation factor of 0.025)} causing \mcmp to not find the optimal solution homotopy. \pump is able to find the best homotopy at all CPs. Note that \mcmp fails to find a solution at a CP of 0.1\% due to not identifying the narrow, short gap before the wide, long passageway is closed. 
(\ref{fig:mcmp_gapSetup}, \ref{fig:mcmp_gapInflation}, \ref{fig:mcmp_gapCost}) A gap in the CP causes \mcmp to be too conservative through the open region at low CPs, while \pump successfully identifies this solution.}
\label{fig:mcmpCompare}
\end{figure}

We begin with a discussion of Monte Carlo Motion Planning (\mcmp) \cite{JansonSchmerlingEtAl2015b},
a recent approach to solving the SKP problem, and show that \pump identifies significantly lower cost solutions in a pair of illustrative examples. 
\mcmp addresses the SKP problem by performing a bisection search over a safety-buffer heuristic correlated with CP, at each step solving a deterministic planning problem with inflated obstacles (higher inflation correlates with lower CP). 
While \cite{JansonSchmerlingEtAl2015b} argues that the algorithm works well in many cases, the assumption that safety-buffer distance is a good analogue for CP breaks down in some planning problems. 
We show two such examples in Figs. \ref{fig:mcmp_degradeSetup} and \ref{fig:mcmp_gapSetup} and show that \pump's $\widehat{\texttt{CP}}$-guided multiobjective search outperforms \mcmp.

Figure~\ref{fig:mcmp_degradeSetup} displays a planning problem in which \revv{the CP of the solution trajectory for the deterministic problem with inflated obstacles has an increase as inflation factor increases (via the discontinuous jump shown in Fig.~\ref{fig:mcmp_degradeInflation})}.
This is a consequence of a narrow, but short gap allowing safer trajectories than a wide, but long passage. 
When iterating on inflation factor, the \mcmp algorithm is unable to locate the solution homotopy class through the narrow gap, as the inflated obstacles block it.
Figure~\ref{fig:mcmp_degradeCost} shows the cost of solutions for various target CPs as compared to \pump, which identifies the narrow gap solution for all CPs. 
For CPs below 3\%, the narrow gap is never identified and instead a more circuitous motion plan is obtained. 
\mcmp's performance is particularly troublesome with a 0.1\% target CP, in which case \mcmp is unable to find any solution as the long passage homotopy class is infeasible, and it never realizes the narrow gap homotopy class. 

The second planning problem, shown in Fig.~\ref{fig:mcmp_gapSetup}, admits two solution homotopies: a narrow passage and a large open region. Figure~\ref{fig:mcmp_gapInflation} shows that as the inflation factor is varied, there is a large discontinuity in CP when the narrow gap becomes blocked. Thus, if the targeted CP is within this gap, the algorithm will converge to exactly the inflation factor that minimally blocks the gap, alternating between motion plans through the narrow passage with high CPs and overly conservative motion plans through the open region, not identifying the desired aggressive motion plan through the open region. Figure~\ref{fig:mcmp_gapCost} shows that \pump is able to find this solution through the open region (unless the gap homotopy is valid for a target CP) while \mcmp is overly conservative, especially at low target CPs. A homotopy blocking heuristic is suggested in \cite{JansonSchmerlingEtAl2015b} for remedying this issue, but this is difficult to execute for planning in general 3D workspaces. 

\subsection{Real-Time Performance and Homotopy Identification}\label{sec:results}

Through the two planning problems in Fig.~\ref{fig:pump_homotopy}, we demonstrate \pump's ability to identify multiple solution homotopy classes and select low-cost trajectories subject to a certified CP constraint, all in a tempo compatible with real-time application (\texttildelow 100 ms). 
The collision probability constraints considered herein are single digit percentages and below, as has been studied for realistic stochastic planning problems in the literature \cite{SunPatilEtAl2015}. 
\revv{The obstacles in our simulations are represented by unions of axis-aligned bounding boxes, as commonly used for broad phase collision detection \cite{LaValle2006}. 
This can provide increasingly accurate representations of the true obstacle set as more are used (e.g., as in Fig.~\ref{fig:pump_homotopy2paths}, or with octree-based environment representations), and accelerates the computation of the local convex regions. 
If needed, more detailed collision checking can be performed during the final smoothing and MC certification phase.}

Table~\ref{tab:timing} details the resulting computation times, solution CPs, and number of partial plans considered, over a range of target CPs and graph sample counts.
Table~\ref{tab:timing} also presents results for a simple repeated RRT method for generating candidate plans. For this comparison, we compute 1000 kinodynamic RRT solution trajectories (as performed in \cite{BergAbbeelEtAl2011,SunPatilEtAl2015}) and select the trajectory with lowest cost subject to MC certification of chance constraint satisfaction.
In all cases \pump run times are on the order of 100 ms, while considering order one hundred thousand partial trajectories in the multiobjective search. 
Note that these solution times are two orders of magnitude faster than those reported for \mcmp \cite{JansonSchmerlingEtAl2015b}, owing to the parallel algorithm design and implementation on GPU hardware.
On average, the graph building accounted for 17\% of the total run time (including computing half-spaces), the exploration for 79\% (55\% overall for \hsmc), and only 4\% for the final trajectory selection, smoothing, and certification phase. \rev{The time for the computation and caching of neighbors and edges is not included, as these computations are performed offline, reducing to constant-time memory access operations at runtime.}
\rev{The initial graph building is very dependent on number and location of obstacles (as this removes invalid samples and edges, and computes the half-spaces for each waypoint), however even with the indoor environment it is dominated in run time by exploration.}
\revv{We further note that for particularly cluttered environments, workspace partitioning schemes (e.g., kd-trees) may be employed to limit scaling.}
\rev{The exploration phase also has some dependence on obstacles through the complexity of the local convex region, however limiting the fidelity of the half-space approximation (number of half-spaces) may keep this scaling in check.}
\revv{Comparing results in Table \ref{tab:timing}, shows that more than number of obstacles, the key driver of computation time is the number of partials plans, which is more dictated by obstacle locations (i.e., difficulty in finding a solution plan) than obstacle count.}
\revv{In fact, comparing only by number of partial plans shows that despite Fig. \ref{fig:pump_homotopyIndoor} having more than five times the number of obstacles of Fig. \ref{fig:pump_homotopy3obs}, the computation time increases by less than a factor of two.}

\begin{table*}[h]
\footnotesize
\centering
\captionsetup{justification=centering}
\caption{\pump vs. repeated RRT (1000 trials) double integrator experimental results.}
\makebox[\textwidth][c]{
\begin{tabular}{| l | c c c | c c c | c c c | c c c | c c c |}

\cline{2-16}
\multicolumn{1}{c |}{} &
    \multicolumn{9}{c |}{Figure~\ref{fig:pump_homotopy3obs}} & 
    \multicolumn{6}{c |}{Figure~\ref{fig:pump_homotopyIndoor}-\ref{fig:pump_homotopy2paths}}\\
\hline
Target CP (\%) & & 0.1 & & & 1 & & & 5 & & & 2 & & & 5 & \\
\hline
Sample Count & 4k & 6k & 8k & 4k & 6k & 8k & 4k & 6k & 8k & 2k & 3k & 4k & 2k & 3k & 4k \\ 
Solution CP (\%) & 0.10 & 0.10 & 0.10 & 1.00 & 1.00 & 1.00 & 5.00 & 5.00 & 5.00 & 1.95 & 1.95 & 1.90 & 4.98 & 4.93 & 4.93 \\
Cost & 2.11 & 2.11 & 2.11 & 2.02 & 2.02 & 2.08 & 1.97 & 1.97 & 1.88 & 3.41 & 2.41 & 1.85 & 1.48 & 1.31 & 1.31 \\
Time (ms) & \multicolumn{1}{c}{48} & \multicolumn{1}{c}{122} & \multicolumn{1}{c |}{216} & \multicolumn{1}{c}{82} & \multicolumn{1}{c}{183} & \multicolumn{1}{c |}{323} & \multicolumn{1}{c}{79} & \multicolumn{1}{c}{223} & \multicolumn{1}{c |}{261} & \multicolumn{1}{c}{97} & \multicolumn{1}{c}{253} & \multicolumn{1}{c |}{282} & \multicolumn{1}{c}{99} & \multicolumn{1}{c}{224} & \multicolumn{1}{c |}{384} \\

$\quad$ \textit{\scriptsize Build Graph Time} & \multicolumn{1}{r}{\scriptsize \textit{13}} & \multicolumn{1}{r}{\scriptsize \textit{28}} & \multicolumn{1}{r |}{\scriptsize \textit{50}} & \multicolumn{1}{r}{\scriptsize \textit{13}} & \multicolumn{1}{r}{\scriptsize \textit{28}} & \multicolumn{1}{r |}{\scriptsize \textit{46}} & \multicolumn{1}{r}{\scriptsize \textit{13}} & \multicolumn{1}{r}{\scriptsize \textit{28}} & \multicolumn{1}{r |}{\scriptsize \textit{49}} & \multicolumn{1}{r}{\scriptsize \textit{10}} & \multicolumn{1}{r}{\scriptsize \textit{25}} & \multicolumn{1}{r |}{\scriptsize \textit{44}} & \multicolumn{1}{r}{\scriptsize \textit{10}} & \multicolumn{1}{r}{\scriptsize \textit{25}} & \multicolumn{1}{r |}{\scriptsize \textit{44}} \\

$\quad$ \textit{\scriptsize Explore Time} & \multicolumn{1}{r}{\scriptsize \textit{33}} & \multicolumn{1}{r}{\scriptsize \textit{92}} & \multicolumn{1}{r |}{\scriptsize \textit{163}} & \multicolumn{1}{r}{\scriptsize \textit{60}} & \multicolumn{1}{r}{\scriptsize \textit{148}} & \multicolumn{1}{r |}{\scriptsize \textit{270}} & \multicolumn{1}{r}{\scriptsize \textit{64}} & \multicolumn{1}{r}{\scriptsize \textit{193}} & \multicolumn{1}{r |}{\scriptsize \textit{206}} & \multicolumn{1}{r}{\scriptsize \textit{68}} & \multicolumn{1}{r}{\scriptsize \textit{205}} & \multicolumn{1}{r |}{\scriptsize \textit{227}} & \multicolumn{1}{r}{\scriptsize \textit{86}} & \multicolumn{1}{r}{\scriptsize \textit{189}} & \multicolumn{1}{r |}{\scriptsize \textit{330}} \\ 

$\quad$ \textit{\scriptsize Selection Time} & \multicolumn{1}{r}{\scriptsize \textit{2}} & \multicolumn{1}{r}{\scriptsize \textit{2}} & \multicolumn{1}{r |}{\scriptsize \textit{3}} & \multicolumn{1}{r}{\scriptsize \textit{9}} & \multicolumn{1}{r}{\scriptsize \textit{7}} & \multicolumn{1}{r |}{\scriptsize \textit{7}} & \multicolumn{1}{r}{\scriptsize \textit{2}} & \multicolumn{1}{r}{\scriptsize \textit{2}} & \multicolumn{1}{r |}{\scriptsize \textit{6}} & \multicolumn{1}{r}{\scriptsize \textit{19}} & \multicolumn{1}{r}{\scriptsize \textit{23}} & \multicolumn{1}{r |}{\scriptsize \textit{11}} & \multicolumn{1}{r}{\scriptsize \textit{3}} & \multicolumn{1}{r}{\scriptsize \textit{10}} & \multicolumn{1}{r |}{\scriptsize \textit{10}} \\ 
Partial Plans & 81k & 288k & 554k & 130k & 367k & 700k & 157k & 570k & 600k & 120k & 284k & 456k & 152k & 331k & 641k \\
\hline
rRRT CP (\%) & \multicolumn{3}{c |}{no satisfactory} & & 0.98 & & & 3.52 & & \multicolumn{3}{c |}{no satisfactory} & \multicolumn{3}{c |}{no satisfactory} \\
rRRT Cost    & \multicolumn{3}{c |}{plan found} & & 2.89 & & & 2.22 & & \multicolumn{3}{c |}{plan found} & \multicolumn{3}{c |}{plan found} \\
\hline
\end{tabular}}\label{tab:timing}
\end{table*}

Figure~\ref{fig:pump_homotopy3obs} shows our first planning problems, in which \pump's multiobjective search phase has identified all important solution homotopies. The little variation in cost with graph resolution (i.e., sample count) demonstrates that \pump generally identifies not only the correct homotopy, but as aggressive a solution as possible within it. Repeated RRT did not encounter a candidate plan with CP below 0.1\%, and achieved significantly worse cost than \pump's principled search for target CPs 1\% and 5\%.

The second planning problem is structured as an indoor flight scenario, once again with multiple solution homotopy classes. Figure~\ref{fig:pump_homotopyIndoor} shows the workspace setup and the Pareto optimal set of trajectories, representing several homotopies, identified by \pump's multiobjective search.
Targeting CPs of 2\% and 5\%, the results shown in Table~\ref{tab:timing} demonstrate that even in this more complex workspace, \pump finds solutions tightly bound on the CP constraint with run times on the order of 100 ms. 
Unlike the previous example, however, a trend of increasing solution quality as graph resolution increases is observed, possibly reflecting the more complex obstacle environment.
Figure~\ref{fig:pump_homotopy2paths} illustrates the cost tradeoff for a 2\% (blue) and 5\% (red) target CP, in which the 2\% motion plan must take a more conservative and costly route (routes visualized through time at \url{https://www.youtube.com/watch?v=ac4A4-ctqrM}).
Repeated RRT did not identify any candidate plans below 5\% CP in this tight hallway environment; the lowest trajectory CP found was 9.57\% with a cost of 1.86. 

\section{Conclusions}\label{sec:conclusions}
In this paper we have introduced the Parallel Uncertainty-aware Multiobjective Planning algorithm for \rev{the stochastic kinodynamic motion  planning problem}. 
\pump represents, to the best of our knowledge, an algorithmic first in that it directly considers both cost and collision probability at equal priority when planning through the free state space. Although multiobjective search is computationally intensive, we have demonstrated that 
\revv{algorithm design and GPU} implementation allow for this principled approach to achieve run times compatible with a \texttildelow10 Hz planning loop. Included in the \pump run time is a Monte Carlo-based certification step to guarantee the collision probability constraint is met, without sacrificing cost metric performance due to conservatism beyond the safety design constraints. 
The Half-Space Monte Carlo method used in our \pump experiments may also be of general interest as an empirically accurate and fast way to approximate motion plan collision probability. 

This work leaves many avenues for further investigation. 
First, we plan to implement this algorithm for more varied dynamical systems and uncertainty models to verify its generality.
Second, we plan to extend this algorithm to environmental uncertainty, particularly in dynamically evolving environments.
Third, we plan to extend the presented multiobjective methodology to other constraints, such as arrival time windows and resource constraints.
Finally, we are in the process of implementing and verifying this algorithm and problem setup onboard a GPU-equipped quadrotor.

\renewcommand{\baselinestretch}{0.91}
\bibliographystyle{IEEEtran-short}
{\footnotesize
\bibliography{../../../bib/main,../../../bib/ASL_papers}
}


\end{document}